\documentclass[10pt,twocolumn,letterpaper]{article}

\usepackage{iccv}              % To produce the CAMERA-READY version
\usepackage{booktabs}
\usepackage{multirow}
\usepackage{caption}
\usepackage{adjustbox}
\usepackage{algorithm}
\usepackage{algpseudocode}
\definecolor{iccvblue}{rgb}{0.21,0.49,0.74}
\usepackage[pagebackref,breaklinks,colorlinks,allcolors=iccvblue]{hyperref}

\def\paperID{12708} % *** Enter the Paper ID here
\def\confName{ICCV}
\def\confYear{2025}

\title{UAVTwin: Neural Digital Twins for UAVs using Gaussian Splatting}

\author{Jaehoon Choi$^{1}$
\and
Dongki Jung$^{1}$
\and
Yonghan Lee$^{1}$
\and
Sungmin Eum$^{2}$
\and
Dinesh Manocha$^{1}$
\and
Heesung Kwon$^{2}$\\
\\
$^{1}$University of Maryland, College Park \:\: $^{2}$DEVCOM Army Research Laboratory
}

\begin{document}
\maketitle

% \footnote{Correspondence to kevchoi@umd.edu}

%%%%%%%%% ABSTRACT
\begin{abstract}
    We present \textit{UAVTwin}, a method for creating digital twins from real-world environments and facilitating data augmentation for training downstream models embedded in unmanned aerial vehicles (UAVs). Specifically, our approach focuses on synthesizing foreground components, such as various human instances in motion within complex scene backgrounds, from UAV perspectives. This is achieved by integrating 3D Gaussian Splatting (3DGS) for reconstructing backgrounds along with controllable synthetic human models that display diverse appearances and actions in multiple poses. To the best of our knowledge, UAVTwin is the first approach for UAV-based perception that is capable of generating high-fidelity digital twins based on 3DGS. The proposed work significantly enhances downstream models through data augmentation for real-world environments with multiple dynamic objects and significant appearance variations—both of which typically introduce artifacts in 3DGS-based modeling. To tackle these challenges, we propose a novel appearance modeling strategy and a mask refinement module to enhance the training of 3D Gaussian Splatting.
    We demonstrate the high quality of neural rendering by achieving a 1.23 dB improvement in PSNR compared to recent methods. Furthermore, we validate the effectiveness of data augmentation by showing a 2.5\% to 13.7\% improvement in mAP for the human detection task.
    % Through extensive experiments, we demonstrate UAVTwin’s effectiveness in enhancing neural rendering quality and boosting the performance of UAV-based perception algorithms. 
    
    % Project Page: \url{https://uavgsim.github.io/}.

\end{abstract}

%%%%%%%%% BODY TEXT
\section{Introduction}
\label{sec:intro}

Unmanned aerial vehicles (UAVs) have become indispensable for human recognition tasks, including person detection \cite{cao2021visdrone,shen2023archangel} and action recognition \cite{barekatain2017okutama,li2021uavhuman,shen2023progressive,shen2024diversifying} in critical applications such as surveillance, disaster response, and security monitoring. Unlike autonomous driving and mobile robots—where extensive, large-scale datasets have propelled rapid progress—UAV-based human identification remains hindered by the scarcity of high-quality datasets. 
The challenges of dataset curation in UAV-based real-world scenarios are twofold: (1) UAVs are constrained by limited flight time, sensor diversity, and variability in camera viewpoints, making large-scale real-world data collection labor-intensive and expensive; and (2) human annotation at UAV altitudes is difficult and error-prone, particularly for fine-grained tasks such as pose estimation.

Given these constraints, synthetic data has emerged as a promising alternative for UAV-based perception \cite{wang2021tartanvo,saini2022airpose,bonetto2023synthetic,khose2024skyscenes,rizzoli2023syndrone} and human activity analysis \cite{yang2023synbody,black2023bedlam}. Recent synthetic dataset pipelines \cite{yang2023synbody,black2023bedlam} leverage virtual environments (e.g., Unreal Marketplace) and rendering engines \cite{dosovitskiy2017carla,shah2018airsim,Blender,Unreal} to generate scene backgrounds, synthetic humans, and UAV camera trajectories. 
However, despite advances in realistic rendering, synthetic data suffers from a significant synthetic-to-real domain gap \cite{shen2023archangel,shen2023progressive,yim2024synplay}, largely due to lighting discrepancies, limited texture fidelity, and unrealistic human appearances.

\begin{figure}[t]
    \centering
    \includegraphics[width=0.99\linewidth]{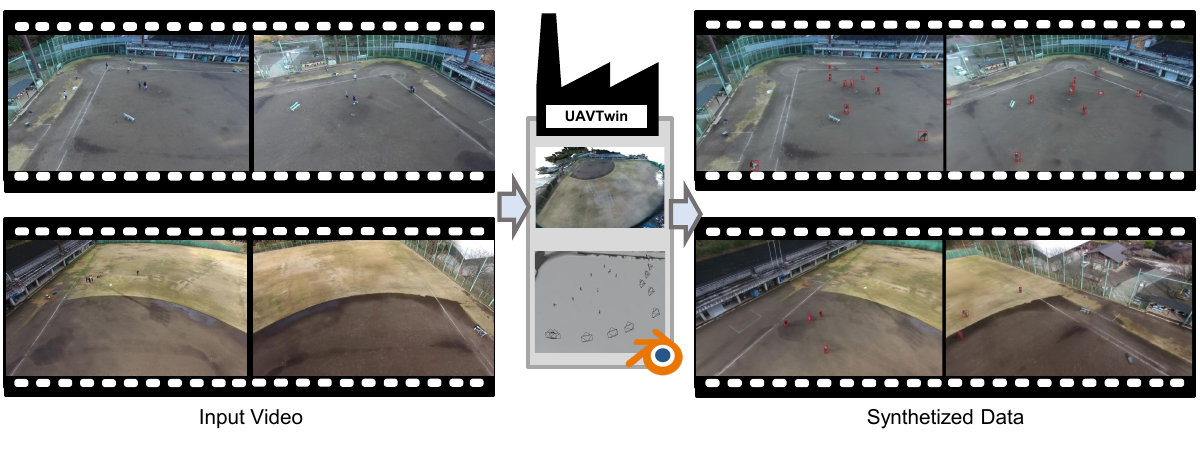}
    \vspace{-3mm}
    \caption{UAVTwin processes video captured by the UAV as input, enabling data generation for training UAV-based human recognition methods.}
    % \vspace{-7mm}
    \label{fig:intro}
\end{figure}

Neural rendering \cite{mildenhall2021nerf,kerbl20233d} has recently emerged as a promising technique to bridge this domain gap, offering photorealistic rendering and digital twin reconstruction of real-world environments. By leveraging 3D reconstruction techniques compatible with rendering engines \cite{Unreal,Blender,dosovitskiy2017carla}, these methods present potential solution for building high-fidelity \textit{digital twin} from real-world data with pixel-accurate annotations. However, \textit{applying neural rendering for UAV-based datasets presents fundamental challenges}:

\begin{itemize}
    \item Appearance Variation Across UAV Captures: UAV-based imagery often exhibits \textit{significant inter-sequence appearance differences}, primarily due to \textit{varying illumination, environmental conditions, and multi-UAV data collection} at different times. It is important to use videos captured across different sensors or times to achieve dense coverage of large areas. While previous methods~\cite{martin2021nerf,chen2022hallucinated,kulhanek2024wildgaussians,zhang2025gaussian} introduce per-image appearance embeddings, these embeddings fail to generalize across unseen UAV sequences and cannot directly transfer to new UAV trajectories. These methods inherently conflicts with our goal of freely generating data for any novel UAV trajectories.
    \item Dynamic Objects and Transient Artifacts: UAV-based scenes frequently contain \textit{moving objects (e.g., vehicles, pedestrians) that occlude the background}, leading to noisy training data. While segmentation models like Segment Anything Model (SAM)~\cite{kirillov2023segment} and its variants~\cite{ren2024grounded} have made progress in 2D scene understanding, they struggle with UAV footage where \textit{dynamic objects are small and sparsely distributed}, occupying fewer pixels than in standard scenes. Additionally, prompt-based segmentation methods~\cite{kirillov2023segment,ren2024grounded} often fail to accurately capture all dynamic objects.
    % , particularly for zero-shot categories.
\end{itemize}

%\noindent\textbf{Main Results:} We introduce \textit{UAVGSim}, a framework that can build a digital twin from complex real-world data captured by UAV and generate novel data for training UAV human-perception algorithms. At the core of our method is a novel integration of 3DGS modeling for building digital twin and data generation using graphics engine. First, we propose Multi-Sequence Gaussian Splatting (MsGS) in \textit{UAVGSim} that incorporates per-sequence appearance embedding for training neural radiance fields for addressing large appearance variation and mask refinement for removing transient and dynamic objects. This model can generate photo-realistic rendering from any novel-view camera trajectory without prior test images and then reconstruct high-fidelity mesh for neural data generation. 

\noindent\textbf{Main Results:} We introduce \textit{UAVTwin}, a new approach that can build a digital twin from real-world data captured by UAV and generate novel data for training UAV-based human recognition algorithms. 
% Our method is an integration of 3DGS modeling for building digital twin and data generation using a graphics engine \cite{Blender}. 
First, we propose Multi-Sequence Gaussian Splatting (MsGS) in \textit{UAVTwin} that incorporates per-sequence appearance embedding for training neural radiance fields for addressing large appearance variation. Additionally, it employs mask refinement to remove transient and dynamic objects. 
Our model can generate photo-realistic rendering from any novel-view camera trajectory without prior test images and then reconstruct a high-fidelity mesh for neural data generation. 
For seamless integration with Blender \cite{Blender}, a 3D mesh that is fully compatible with its framework is required.

Next, we explore a new approach to achieve data augmentation in UAV scenes by integrating MsGS and data generation using a graphics engine \cite{Blender}. 
Our system can import synthetic humans from digital assets and environmental lighting, and generate camera poses that mimic UAV flight trajectory for realistic data generation. 
We can select various motions for synthetic humans and generate various scenarios for data augmentation. 
We further incorporate scene composition for enhancing realism of generated data. Thanks to scene backgrounds rendered by MsGS and composition technique, 
our system can minimize the domain gap between virtual environment and the real world.   

\begin{itemize}[leftmargin=*]
    \item We propose a MsGS for training 3DGS-based neural radiance field, which utilizes per-seqeunce embedding to enable varying appearance modeling from multi-sequence image collections and refine semantic masks for addressing dynamic objects. 
    \item We present a neural data generation approach that integrates the strengths of neural rendering and graphics engines to augment training data for UAV-based perception algorithms.
    \item We validate the background rendering quality, demonstrating that our method outperforms other Gaussian Splatting algorithms by 2.79 dB PSNR on the Okutama-Action dataset. Additionally, we assess the effectiveness of augmented data by training a person detection model, achieving a 2.5\%–13.7\% improvement in mAP.
\end{itemize}

\begin{figure*}[t]
    \centering
    \includegraphics[width=0.99\linewidth]{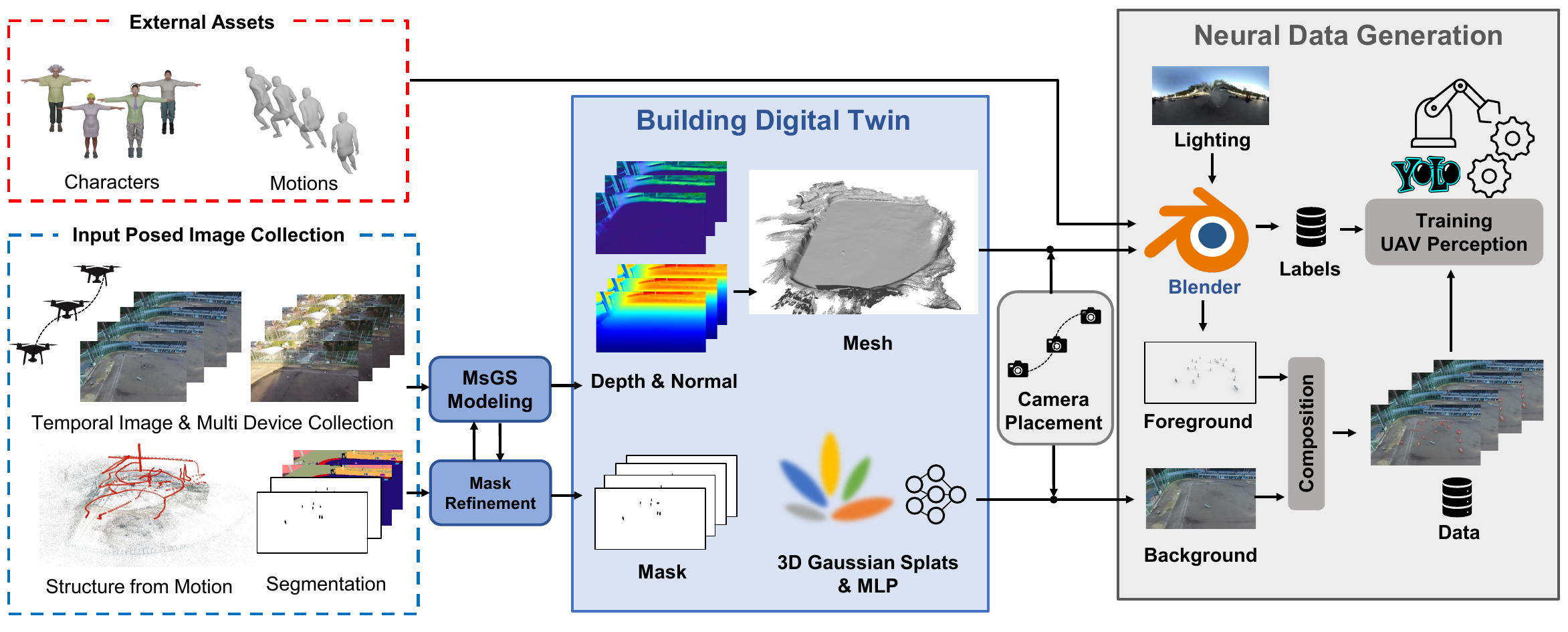}
    \vspace{-4mm}
    \caption{\textbf{UAVTwin framework.} In Section \ref{sec:buildingdigialtwin}, our approach first constructs a digital twin using UAV-based images captured at different times. We introduce MsGS, a novel 3DGS method to analyze varying appearance images and reconstruct a clean mesh, Gaussian splats, and an MLP for novel-view synthesis. Then, in Section \ref{sec:neuraldatageneration}, our method generates data by compositing foreground humans rendered in Blender with backgrounds rendered using trained Gaussian splats.}
    \label{fig:framework}
    % \vspace{-7mm}
\end{figure*}

%%%%%%%%% Related Work
\section{Related Work}
\label{sec:relatedwork}

\noindent\textbf{Neural Rendering.}
Neural rendering includes methods that utilize neural networks to replace or improve traditional rendering pipelines. 
Neural Radiance Fields (NeRF) \cite{mildenhall2021nerf} have demonstrated remarkable photorealistic rendering quality \cite{zhang2020nerf++,fridovich2022plenoxels,chen2022tensorf,muller2022instant,sun2022direct,barron2021mip,barron2022mip,rematas2022urban,barron2023zip} and enabled high-fidelity surface reconstruction \cite{wang2021neus,yariv2021volsdf,oechsle2021unisurf,yu2022monosdf,choi2023tmo,yariv2023bakedsdf,li2023neuralangelo,Choi2024LTM}. 
More recently, 3D Gaussian splatting (3DGS) \cite{kerbl20233d,yu2024mip} has achieved a comparable rendering quality while offering significantly superior rendering speed compared to NeRF. 
However, 3DGS often lacks precise alignment with the underlying scene geometry, negatively affecting rendering quality and surface extraction. 
Some studies \cite{qian2023gaussianavatars,waczynska2024games,choi2024meshgs,lee2024mode} leverage off-the-shelf reconstruction methods and align Gaussian splats with this geometry. 
Several approaches \cite{guedon2023sugar,huang20242d,yu2024gaussian,chen2024pgsr,zhang2024rade,yu2024gsdf} have explored the use of 3D Gaussian splats not only for rendering but also for surface extraction. 
In this work, we present a method to train 3D Gaussian splats with UAV captured images in order to achieve both photorealistic rendering and surface reconstruction. 
% In this work, we build a digital twin using 3D Gaussian splats to achieve both photorealistic rendering and surface reconstruction.  

\noindent\textbf{Data Generation for UAV Perception Tasks.}
Synthetic data with dense annotation is a practical solution to train learning-based UAV perception tasks. 
The combination of Airsim \cite{shah2018airsim} and Unreal engine \cite{Unreal} is widely utilized for various tasks including human pose and shape estimation \cite{saini2022airpose}, visual odometry \cite{wang2021tartanvo}, and animal detection \cite{bonetto2023synthetic}. 
SkyScene \cite{khose2024skyscenes} and SceneDrone \cite{rizzoli2023syndrone} leverage the CARLA simulator \cite{dosovitskiy2017carla} to generate large-scale datasets with dense annotations for tasks such as object detection. 
Several works \cite{shen2023archangel,reddy2023synthetic,yim2024synplay} utilize Unity to generate data for aerial-view action recognition and human detection. However, these methods continue to highlight the significant domain gap between real and synthetic data. 
UAV-Sim \cite{maxey2024uav} is limited to novel-view synthesis and cannot augment new data beyond this, as it lacks the ability to control real humans from captured data. Thus, our goal is to develop a photorealistic digital twin for UAV perception tasks.

\noindent\textbf{Gaussian Splatting for Simulation and Robotics.}
Recent innovations in neural rendering methods enable building digital twin from the real-world environment. Many robotics applications \cite{irshad2024neural,wang2024nerf} leverage this photorealistic simulator to learn the real-world and generate the data for specific application. 
Robotics manipulation studies \cite{meyer2024pegasus,qureshi2024splatsim,wu2024rl,lou2024robo,han2025re} integrate a physics engine with Gaussian splatting (GS) rendering and reconstruct high-fidelity objects with physical properties, enabling the generation of photorealistic manipulation data. 
VR-Robo \cite{zhu2025vr} enables photorealistic simulation for robot navigation and locomotion. 
Due to the nature of these tasks, achieving photorealistic rendering is relatively straightforward, as data can be easily collected in a controlled laboratory environment with consistent background and illumination. 
For UAV applications, previous studies \cite{chen2024splat,quach2024gaussian,low2024sous} have developed GS-based simulators using data collected in controlled laboratory environments, enabling end-to-end training for UAV navigation. 
However, to the best of our knowledge, our work is the first GS-based digital twin generation method designed for training UAV perception algorithms, with a particular focus on human-related tasks such as person detection.

% \noindent\textbf{Human Data Agumentation}
% On the Equivalency, Substitutability, and Flexibility of Synthetic Data
% Pointodyssey, Xrfeitoria,

\section{Building Digital Twin}
\label{sec:buildingdigialtwin}

The \textit{UAVTwin} system takes UAV-captured images as input and constructs a digital twin to generate data for training UAV perception algorithms. 
An overview of our pipeline is illustrated in Fig \ref{fig:framework}. 
The UAV collects $N$ video sequences \textbraceleft$\overline{V}_{j}$\textbraceright$_{j=1}^{N}$ from the real-world environments. 
These sequences are captured at different times, from varying UAV trajectories and altitudes, and under diverse lighting conditions. Each video sequence, denoted as $\overline{V_{j}} = $\textbraceleft$I_{i}$\textbraceright$_{i=1}^{K_{j}}$, consists of $K_{j}$ consecutive images, 
each exhibiting various characteristics. For all $\mathcal{T}$ images, we extract camera poses and initial point clouds using COLMAP \cite{schonberger2016structure}. 
% We design our method based on the recent 3D Gaussian Splatting (3DGS) \cite{kerbl20233d} to reconstruct accurate 3D geometry for background mesh and enable photorealistic rendering for background region, which are basic component of rendering engine. 
We propose Multi-sequence Gaussian Splatting (MsGS) to reconstruct accurate 3D geometry for the background mesh and enable photorealistic rendering for the background region, which are basic components of rendering engine. 
To handle transient objects, two types of segmentation maps are extracted from all images using both the SAM \cite{kirillov2023segment,ren2024grounded} and entity segmentation methods \cite{qilu2023high}. Both segmentation maps are used for training the 3D Gaussian model and refining mask quality.

\begin{figure*}[t]
    \centering
    \includegraphics[width=0.99\linewidth]{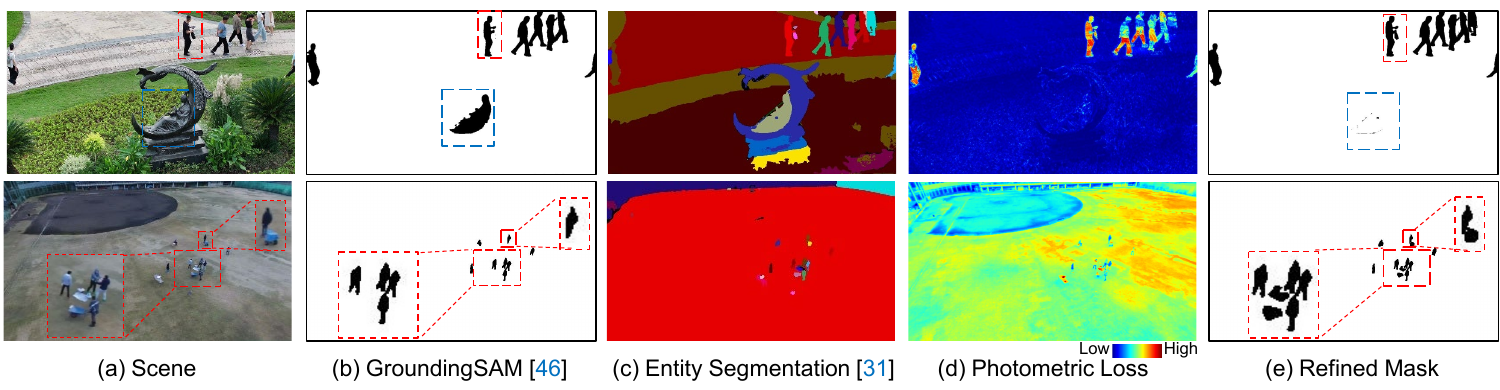}
    \vspace{-2mm}
    \caption{\textbf{Mask Refinement.} (a) is an example of training images with dynamic objects. (b) is its segmentation masks $\overline{M}$ using GroundingSAM \cite{ren2024grounded}. (c) is the entity segmentation masks $\hat{M}$ \cite{otonari2024entity}. Based on the SAM masks in (b), we add entity masks with high photometric loss (red dotted boxes) and remove those with low photometric loss (blue dotted boxes), resulting in the refined masks shown in (e).}
    \label{fig:segmentation}
    % \vspace{-3mm}
\end{figure*}

\begin{figure}[t]
    \centering
    \includegraphics[width=0.95\linewidth]{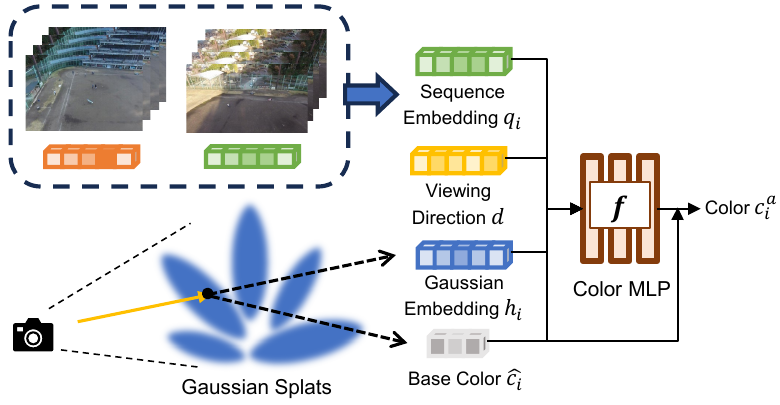}
    \vspace{-2mm}
    \caption{\textbf{The core components of MsGS.} From a video sequence captured at different times with varying appearances, we define a sequence embedding $q_{i}$.  For novel-view image rendering, we select a specific sequence.  Our color MLP $f$ takes as input the sequence embedding, viewing direction embedding, per-Gaussian embedding, and base color. The output color is modulated to account for appearance variations.}
    \label{fig:gsmlp}
    % \vspace{-3mm}
\end{figure}

\subsection{Preliminaries: 3D Gaussian Splatting}
\label{sec:3.1}

3D Gaussian Splatting (3DGS) \cite{kerbl20233d} represents a 3D scene using a set of anisotropic 3D Gaussians $G$. Each Gaussian incorporates multiple attributes to represent both the geometry and appearance of the scene.
Each 3D Gaussian is parameterized using its mean position $\mu_{i} \in \mathbb{R}^3$, and 3D covariance matrix $\Sigma_{i} \in \mathbb{R}^{3\text{x}3}$ in world space.
% \begin{equation} \label{eq:3DGaussians}
% G_{i}(x) = exp(-\frac{1}{2}(x-\mu_{i})^T \Sigma_{i}^{-1}(x-\mu_{i})).
% \end{equation}
During optimization, the covariance matrix is reparametrized using a scaling matrix $S_{i} \in \mathbb{R}^{3}$ and a rotation matrix $R_{i} \in \mathbb{R}^{3\text{x}3}$, which is represented by a quaternion $q \in \mathbb{R}^4$, following the formulation $\Sigma_{i} = R_{i}S_{i}{S_{i}}^{T}{R_{i}}^{T}$. 
$P_{CW}$ is the world-to-camera transformation matrix and $W$ is the rotational component of $P_{CW}$. 
For rendering, the 3D Gaussians are transformed into the camera coordinates with $W$ and projected to the image space using the affine transformation matrix $J_{i}$ following \cite{zwicker2002ewa}. 
The 2D Gaussians $g_{i}(\mu'_{i}, \Sigma'_{i})$ on the image plane is represented as $\mu'_{i} = \pi(P_{CW} \cdot \mu_{i})$ and $\Sigma'_{i} = J_{i}W\Sigma_{i}W^{T}{J_{i}}^{T}$.  
% \begin{equation} \label{eq:3DGaussianProjection}
% \mu'_{i} = \pi(P_{CW} \cdot \mu_{i}), \:\:\: \Sigma'_{i} = J_{i}W\Sigma_{i}W^{T}{J_{i}}^{T}
% \end{equation}
$\pi$ is the projection operation and $\Sigma'$ is the 2D covariance matrix in the image space. 
3DGS sorts all the Gaussian involved in a pixel and renders a 2D image via $\alpha$-blending using the following equation:
\begin{equation} \label{eq:3DGSrendering}
\bar{C} = \sum_{i=1}^{n} c_{i}\alpha_{i}\prod_{j=1}^{i-1}(1-\alpha_{j}), \:\:\:\: \alpha_{i} = o_{i}g_{i}(x).
\end{equation}
where $c_{i}$ and $\alpha_{i}$ represent the color and opacity of this point multiplied by an optimizable per-point opacity and SH color coefficients. 
These 3D Gaussians are optimized with respect to the photometric loss $L_{pho}$ using a combination of SSIM \cite{wang2004image} and $L_{1}$ losses between the rendered color $\bar{C}$ and groundtruth colors $C$: 
\begin{equation} \label{eq:rgbloss}
    L_{pho} = (1-\lambda_{pho})\text{L1}(\bar{C},C) + \lambda_{pho}\text{SSIM}(\bar{C}, C)).
\end{equation}

\subsection{Multi-sequence Gaussian Splatting (MsGS)}
\label{sec:3.2}
Previous research commonly employs either per-image embeddings \cite{martin2021nerf,rematas2022urban,kulhanek2024wildgaussians} or Convolutional Neural Network (CNN)-based appearance encoders \cite{chen2022hallucinated,zhang2025gaussian} to address appearance variations. 
However, these approaches are not well-suited for our data generation pipeline, as it is difficult to accurately capture the appearance of unseen images produced by a random UAV trajectory. 
The proposed MsGS uses per-sequence embedding \textbraceleft$q_{i}$\textbraceright$_{i=1}^{N}$ for each sequence.
% The original 3DGS \cite{kerbl20233d} utilizes spherical harmonics (SH), which struggle to encode large appearance variations due to their explicit color representations. 
The original 3DGS \cite{kerbl20233d} utilizes spherical harmonics (SH), which may not work well in terms of encoding large appearance variations due to their explicit color representations. 
Instead, we use 3-layer color MLP $f$ of width 128. 
As shown in Fig. \ref{fig:gsmlp}, we define a color MLP $f$ that takes the viewing direction embedding $d$, per-sequence embedding $q_{i}$, base color $\hat{c_{i}}$, and a learnable embedding for each Gaussian $h_{i}$, as input. Following the previous works \cite{rematas2022urban,kulhanek2024wildgaussians,chen2024pgsr}, the color MLP is designed to output the parameters of an affine transformation, $\alpha$ and $\beta$.
\begin{equation} \label{eq:AppearanceMoedling}
(\alpha_{i}, \beta_{i}) = f(h_{i}, q_{i}, d, \hat{c_{i}}), \:\:\: c^{a}_{i} = \alpha_{i} \hat{c_{i}} + \beta_{i},
\end{equation}
where $c^{a}_{i}$ represents the toned color corresponding to the assigned sequence. 
Then, the toned color for each Gaussian splat is passed through the 3DGS rasterization process to render the final image map $C^{a}$ for each camera view. 
The benefit of per-sequence embedding is that we can freely generate data for novel-view poses without utilizing prior test-view images like previous research \cite{kulhanek2024wildgaussians,zhang2025gaussian}. 
Furthermore, we enforce consistent appearance within the same sequence and utilizes per-sequence embeddings to capture variations in appearance across different sequences. 
Additionally, incorporating a per-sequence embedding is effective in mitigating floaters, improving not only rendering quality but also surface reconstruction. 

\subsection{Mask Refinement}
\label{sec:3.3}
To mitigate the influence of transient objects, our approach requires accurate masks for moving objects. 
Previous uncertainty models based on DINO \cite{kulhanek2024wildgaussians} are ineffective due to the extremely small size of these moving objects. 
Instead, given $\mathcal{T}$ images, we use GroundingSAM \cite{ren2024grounded} to extract image masks. 
To detect bounding boxes associated with specific text phrases, we provide a text caption (e.g., ``human") as input for text embedding in Grounding DINO \cite{liu2024grounding}. 
Using these box prompts, SAM \cite{kirillov2023segment} generates mask annotations $\overline{M} = $\textbraceleft${M_{i}}$\textbraceright$_{i=1}^{\mathcal{T}}$. However, GroundingSAM often produces inaccurate masks.     

To address this, we introduce a novel mask refinement module to refine the accuracy of the SAM masks $\overline{M}$ in Fig.\ref{fig:segmentation}-(a). 
Inspired by Entity-NeRF \cite{otonari2024entity}, we refine masks leveraging the entity masks \textbraceleft${\hat{M}_{i}}$\textbraceright$_{i=1}^{\mathcal{T}}$  in Fig.\ref{fig:segmentation}-(b) extracted through entity segmentation \cite{qilu2023high}.  
We first measure error maps \textbraceleft${R_{i}}$\textbraceright$_{i=1}^{\mathcal{T}}$ using photometric loss $L_{pho}$ in Fig.\ref{fig:segmentation}-(c) in Eq. \ref{eq:rgbloss} for each training views. 
For a set of pixels $S(e_{id})$ corresponding to an entity ID $E_{id}$, we compute error map $R(e_{id})$ per each entity as follows,
\begin{equation} \label{eq:errormap}
R(e_{id}) = \frac{\sum_{i \in S(e_{id})}L_{pho}(i)}{A(S(e_{id}))}
\end{equation}
where $A(\cdot)$ represents the area of the corresponding pixels. 

Here, we make three assumptions: (1) in UAV capture scenarios, most dynamic objects occupy a small number of pixels, (2) transient objects exhibit significant RGB loss errors, and (3) if a cluster of pixels in SAM masks has an RGB loss error below a certain threshold, it can be considered a static mask. 
Thus, we remove entity masks and SAM masks that exceed the area thresholds $\rho_1$ and $\rho_2$ respectively and select entity masks $S'(e_{id})$ where $A(S(e_{id})) < \rho_1$ and $A(\overline{M}) < \rho_2$.
Next, we identify all entity masks that overlap with the dilated SAM masks $\overline{M}$ and gather these overlapped masks $M' \gets Dilate(\overline{M}) \cup \hat{M}$.  
To determine whether an entity mask should be added to or removed from the original SAM masks $\overline{M}$, we apply an RGB loss threshold $\rho_{rgb}$. 
For each entity mask $\hat{M}(e_{id}) \in M'$, we include $\hat{M}(e_{id})$ to $\overline{M}$ if $R(e_{id})$ exceeds $\rho_{pho}$; otherwise, we remove $\hat{M}(e_{id})$ from $\overline{M}$.    
We outline the pseudo code for our mask refinement in the supplementary material. 

\subsection{Geometry Reconstruction}
\label{sec:3.4}
Following PGSR \cite{chen2024pgsr}, we first render the normal maps $N$ and distance map $\hat{D}$ of the plane and then convert them into depth maps $D$ using camera intrinsic matrix $K$.
\begin{equation} \label{eq:renderingdepth}
D(p) = \frac{\hat{D}}{N(p)K^{-1}\widetilde{p}},
\end{equation}
where $p\in \mathbb{R}^2$ is the pixel coordinate and $\hat{p}$ is the homogeneous coordinate of $p$. Furthermore, we apply the scale regularization to flatten the 3D Gaussian splats. Given the scale parameters $S_{i} = (s_1, s_2, s_3)$, we minimize the minimum scale as follows:

\begin{equation} \label{eq:scaleloss}
L_{scale} = \sum_{i \in G}(\lambda_{s}|\text{min}(s_{1}, s_{2}, s_{3})|).
\end{equation}

\subsection{Training Objectives for MsGS}
\label{sec:3.5}
Our training process consists of two stages. In the first stage, we train 3D Gaussian attributes and MLP $f$ using the original SAM masks $\overline{M}$, applying both scale regularization in Eq. \ref{eq:scaleloss} and masked photometric loss $L_{Mpho}$. The masked photometric loss is defined as follows: 
\begin{equation} \label{eq:masked_rgb_loss}
    L_{Mpho} = (1-\lambda_{pho})\overline{M}\text{L1}(C^{a},C) + \lambda_{pho}\overline{M}\text{SSIM}(C^{a}, C)),
\end{equation}
where the mask $\overline{M}$ is a binary mask and is multiplied by the per-pixel loss from Eq. \ref{eq:rgbloss}. $C^{a}$ is the rendered color of the rasterized images.

In the second stage, we apply our mask refinement technique from Sec. \ref{sec:3.3} to enhance the original SAM masks $\overline{M}$, resulting in a high-quality mask $\widetilde{M}$. Additionally, we apply masked single-view normal loss $L_{Msvgeo}$ and multi-view regularization $L_{mvreg}$ from previous studies \cite{chen2024pgsr,yu2024gaussian,campbell2008using}. The details of these two losses are provided in the supplementary materials. The overall loss function $L$ is defined as:
\begin{equation} \label{eq:total_loss}
    L = L_{Mpho} + L_{scale} + L_{Msvgeo} + L_{mvreg}.
\end{equation}

\subsection{Mesh Reconstruction}
\label{sec:3.6}
We first render the depth for each training view and then apply the TSDF Fusion algorithm \cite{newcombe2011kinectfusion} to construct the corresponding TSDF field. Subsequently, we extract the mesh $B$ for the background region. It is essential that Gaussian splats are precisely aligned with the actual 3D geometry represented by mesh $B$ to ensure that the background image rendered by the Gaussian splats can be seamlessly composited with the foreground humans, which are positioned relative to the mesh.

\section{Neural Data Generation}
\label{sec:neuraldatageneration}

For data generation, our approach comprises synthetic human placement, camera trajectory generation, and scene composition. Our camera trajectory generation and synthetic human placement are implemented based on the reconstructed mesh $B$ from Section \ref{sec:3.6}. 
This ensures seamless composition of the foreground humans, rendered using Blender \cite{Blender}, and the background, rendered with our neural rendering method in Section \ref{sec:3.2}, providing data for UAV perception.

\subsection{Camera Trajectory Generation}
The camera trajectory is a critical factor in generating UAV data. 
We consider the camera's pitch angle and various camera movement patterns for data collection. 
Following Blender convention, we define the camera's location $t$ and rotation $r$ to generate the camera trajectory. To simulate a real drone's movement, we introduce Gaussian random noise $\epsilon^{r}, \epsilon^{t} \in \mathbb{R}^3$ to each components, where the camera' location is represented as $(t_{x} + \epsilon^{t}_{x}, t_{y} + \epsilon^{t}_{y}, t_{z} + \epsilon^{t}_{z})$ and its rotation as $(r_{x} + \epsilon^{r}_{x}, r_{y} + \epsilon^{r}_{y}, r_{z} + \epsilon^{r}_{z})$.    

\noindent\textbf{Translational Trajectory}: The UAV moves horizontally or vertically by increasing or decreasing $t_{x}, t_{y}$ in a straight line while maintaining a fixed camera orientation, independent of the subject.

\noindent\textbf{Stationary Yaw Rotation}: The UAV remains in place but rotates horizontally $r_{z} \in [0, 2\pi]$ to capture a panoramic view of the environment.

\noindent\textbf{Orbit Trajectory}: The UAV moves in a circular path around the subject, maintaining a certain distance while keeping the camera focused on it. We compute the center points $p^{c}$ of averaging all actors and define azimuthal angle $\phi$ and radius $r_{a}$. This camera is parameterized by $(r_{a}cos(\phi)+p^{c}_{x}, r_{a}sin(\phi)+p^{c}_{y},t_{z}+p^{c}_{z})$ using a human-centric spherical system.

\noindent\textbf{Altitude-Varying Trajectory}: The UAV dynamically adjusts its altitude by increasing or decreasing $t_{z}$, ascending to expand the field of view for broader coverage and descending to capture finer details with focused observation.

\begin{figure}[t]
    \centering
    \includegraphics[width=0.99\linewidth]{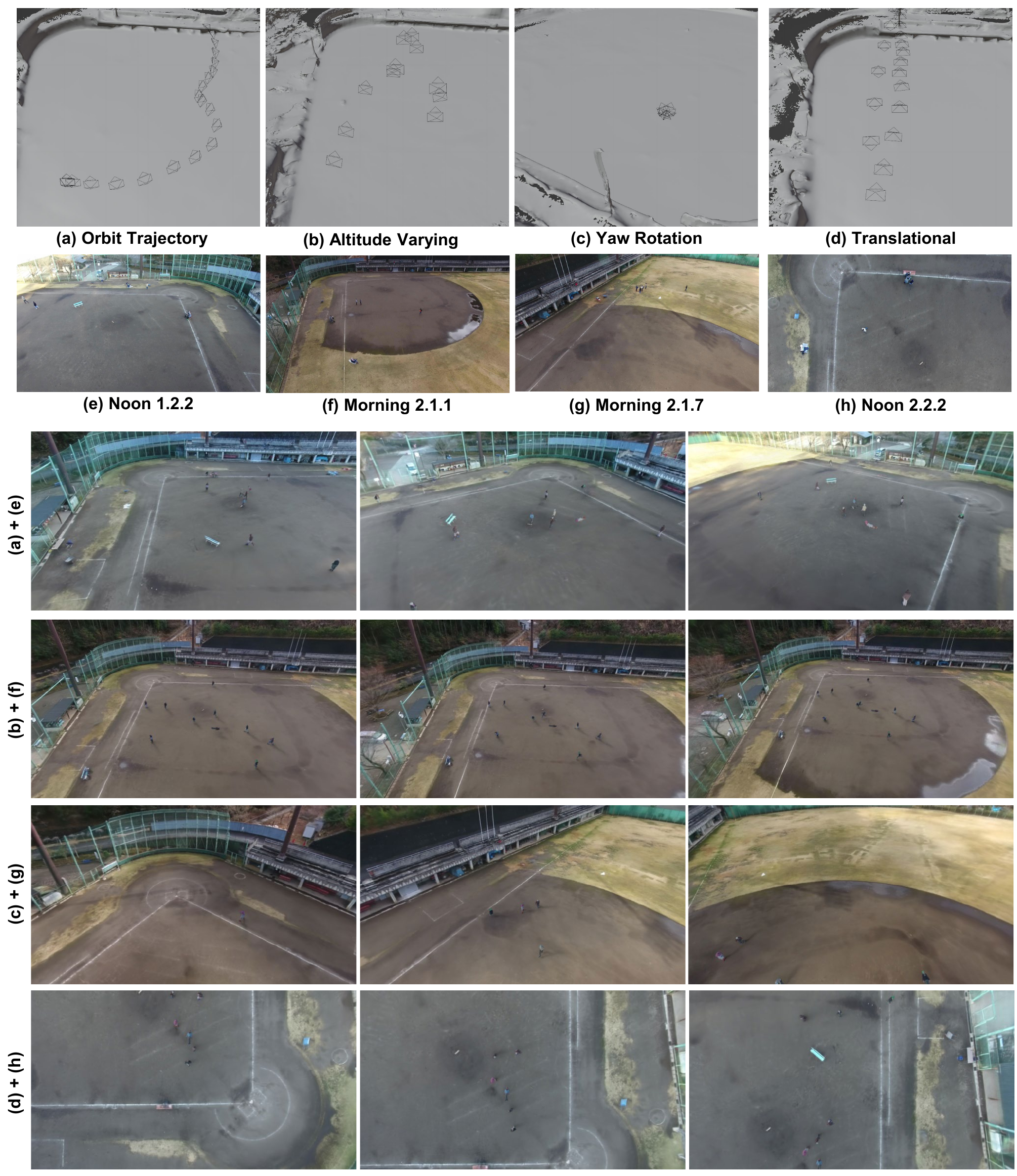}
    \vspace{-2mm}
    \caption{\textbf{Qualitative Results of Synthetized Data.} The first row illustrates the camera trajectory, while the second row presents a sample training image from the sequence used as a per-sequence embedding input for rendering our MsGS. Data is generated using different trajectories with sequence embeddings extracted from training data: (a)+(e) orbit (Noon 1.2.2), (b)+(f) altitude varying (Morning 2.1.1), (c)+(g) yaw rotation (Morning 2.1.7), and (d)+(h) translational (Noon 2.2.2).}
    % \vspace{-6mm}
    \label{fig:data_generation_result}
\end{figure}

\subsection{Synthetic Human Placement}
After generating the camera trajectory in the previous stage, we utilize SynBody \cite{yang2023synbody}, incorporating SMPL-XL models \cite{pavlakos2019expressive} to simulate human actors with diverse body shapes and clothing. 
Using the reconstructed mesh from Section \ref{sec:3.4}, we identify the mesh faces visible from all camera views along the given trajectory. 
This allows us to determine the feasible range for synthetic human placement. 
Within this range, we randomly partition the area for actor placement and position $N_{a}$ actors. 
For each human actor, we randomly select a single-person motion from AMASS \cite{mahmood2019amass}, excluding interactive or non-ground motions. 

% \textcolor{red}{human scale!!!}

%%%%%%%% Okutama-Action
\begin{table*}[h]
    \centering
    \begin{adjustbox}{width=0.95\linewidth,center}
    \begin{tabular}{l|c|c|ccccc}
    \toprule
    \multirow{2}{*}{Method}&\multirow{2}{*}{Mesh}&\multirow{2}{*}{Mask}&\multicolumn{4}{c}{PSNR \(\uparrow\)  / SSIM \(\uparrow\)  / LPIPS \(\downarrow\) }\\
    & & & Drone1-Morning & Drone2-Morning & Drone1-Noon & Drone2-Noon & Mean \\
    \midrule
    3DGS \cite{kerbl20233d} & X & O &  26.33 / 0.901 / 0.222  &  27.32 / 0.891 / 0.216 & 31.47 / 0.917 / 0.177 & 34.10 / 0.942 / 0.175 &  29.90 / 0.913 / 0.198  \\
    % 3DGS \cite{yu2024mip} & X & O  & 0.00 / 0.00 / 0.00 & 0.00 / 0.00 / 0.00 & 0.00 / 0.00 / 0.00 & 00.0 / 00.0 / 00.0 \\
    % mip-GS \cite{yu2024mip} & O & X & 0.00 / 0.00 / 0.00 & 0.00 / 0.00 / 0.00 & 0.00 / 0.00 / 0.00 & 00.0 / 0.00 / 0.00 \\
    % \midrule
    GOF \cite{yu2024gaussian} & O & O & 29.04 / 0.918 / 0.164 & 26.43 / 0.896 / 0.212 & 29.95 / 0.914 / 0.189 & 32.09 / 0.938 / 0.193 & 29.35 / 0.916 / 0.190 \\ 
    2DGS \cite{huang20242d} & O & X & 27.09 / 0.899 / 0.236 & 28.33 / 0.901 / 0.222 & 31.22 / 0.923 / 0.197 & 32.49 / 0.936 / 0.211 & 30.56 / 0.924 / 0.198  \\ 
    PGSR \cite{chen2024pgsr} & O & X & 25.57 / 0.906 / 0.188 & 26.11 / 0.907 / 0.166 & 29.17 / 0.936 / \underline{0.137} & 31.94 / \textbf{0.950} / \textbf{0.143} & 28.25 / 0.925 / \underline{0.155} \\
    PGSR \cite{chen2024pgsr} & O & O & 27.71 / 0.920 / \textbf{0.142} & 26.96 / 0.911 / \textbf{0.156} & 29.47 / 0.937 / \textbf{0.137} & 31.84 / \underline{0.946} / \underline{0.152} & 29.00 / 0.921 / \textbf{0.147} \\
    \midrule
    MsGS (Ours) & O & O & \textbf{31.07} / \textbf{0.925} / \underline{0.166} & \textbf{29.89} / \textbf{0.918} / \underline{0.161} & \textbf{33.08} / \textbf{0.937} / 0.154 & \underline{33.59} / 0.944 / 0.188 & \textbf{31.79} / \textbf{0.932} / 0.162 \\ 
    MsGS w/o Refine & O & O & \underline{30.43} / 0.920 / 0.176 & \underline{29.83} / \underline{0.918} / 0.162 & \underline{32.96} / \underline{0.937} / 0.161 & \textbf{33.60} / 0.941 / 0.178 & \underline{31.75} / \underline{0.929} / 0.166 \\ 
    MsGS w/o Mask & O & X & 30.32 / \underline{0.921} / 0.168 & 29.63 / 0.916 / 0.162 & 32.64 / 0.935 / 0.160 & 33.46 / 0.945 / 0.168 & 31.71 / 0.929 / 0.171 \\ 
    \bottomrule
    \end{tabular}
    \end{adjustbox}
    \vspace{-2mm}
    \caption{\textbf{Quantitative Comparison on Okutama-Action Dataset.} Except for 3DGS \cite{kerbl20233d}, all methods are designed for surface reconstruction. 
    The ``Mask" column indicates whether a mask is applied when measuring the training loss. Compared to 2DGS \cite{huang20242d} and PGSR \cite{chen2024pgsr}, our method enhances the PSNR by 1.23 dB and 2.79 dB, respectively.
    The best and second best-performing algorithms for each metric are bolded and underlined.
    }
    % \vspace{-3mm}
    \label{table:rendering_quantitative_comparison_okutama}
\end{table*}

\begin{figure*}[h]
    \centering
    \includegraphics[width=0.99\linewidth]{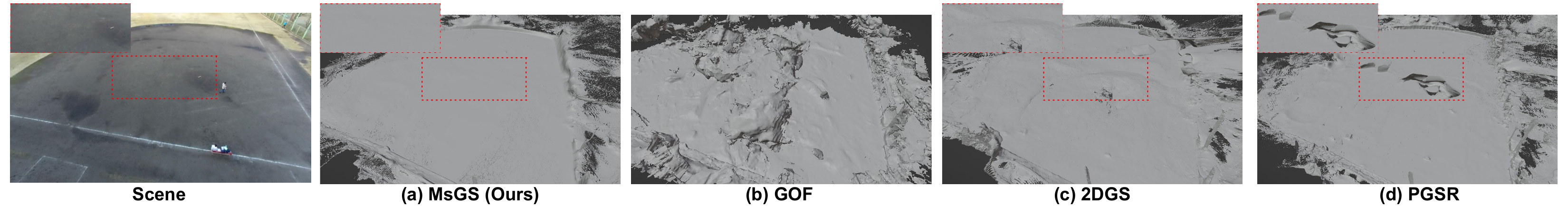}
    \includegraphics[width=0.99\linewidth]{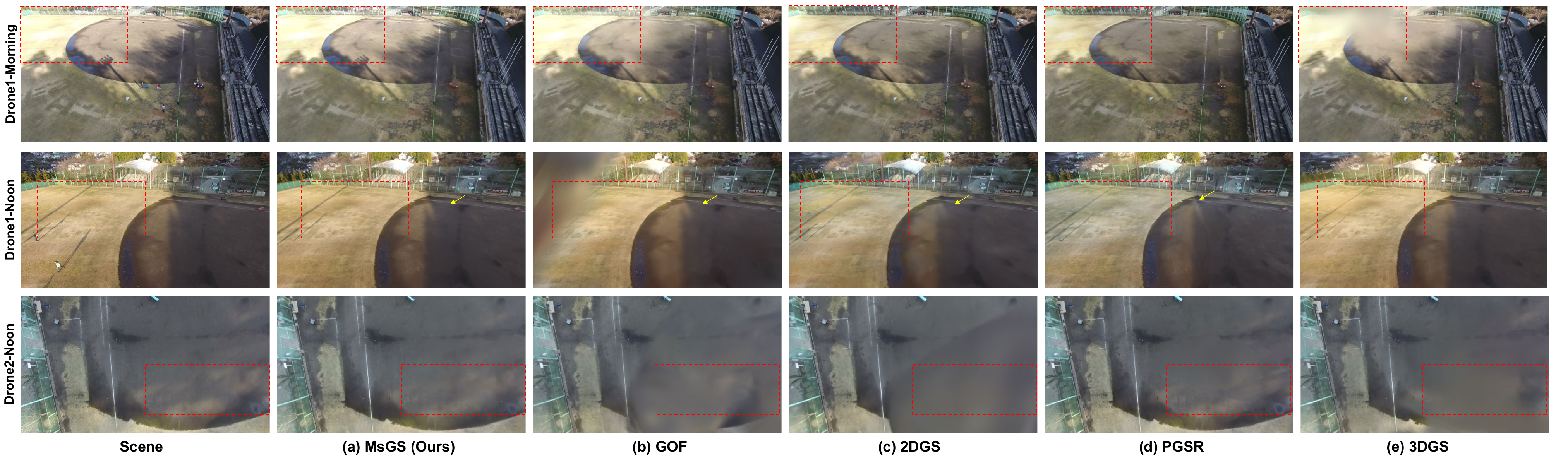}
    \vspace{-4mm}
    \caption{\textbf{Qualitative Comparison of MsGS (Our method), GOF \cite{yu2024gaussian}, 2DGS \cite{huang20242d}, PGSR \cite{chen2024pgsr}, and 3DGS \cite{kerbl20233d}.} In the first row, we show shading mesh without texture. The red dotted lines illustrates the superior geometric quality of our method. We visualize the rendering results (2-4 rows). Red dotted lines and yellow arrow emphasize difference in rendering quality.}
    \label{fig:rendering_surface_comparison}
    % \vspace{-4mm}
\end{figure*}

\subsection{Scene Composition}
\label{sec:4.2}

To achieve high realism for scene composition, we require high dynamic range (HDR) lighting of the target scene to utilize as a virtual light source. 
We employ DiffusionLight \cite{phongthawee2024diffusionlight}, a diffusion-based lighting method, to generate a high dynamic range (HDR) map by inpainting a chrome ball into the image. 
The resulting environment map is then imported into the rendering engine as an environmental light source. 

For foreground rendering, our approach utilizes the Blender's Cycles renderer \cite{Blender} graphics engine to render foreground humans. 
We set the large plane under humans and use the shadow catcher module in Blender to create a shadow effect. 
We alpha-blend this foreground object $I_{fg}$ with background rendered images $I_{bg}$ from our neural renderer in Section \ref{sec:3.2}. 
We also add motion blur $G(\cdot)$ with Gaussian kernel to rendered foreground humans.   
\begin{equation} \label{eq:scene_composition}
I_{comp} = \alpha G(I_{fg}) + (1 - \alpha) I_{bg}
\end{equation}
where $\alpha$ denotes the alpha map rendered by Blender. One key aspect we want to emphasize is that since the graphics engine can generate precise ground truth for segmentation masks, depth maps, and action labels, we can effectively augment the dataset with highly accurate ground truth annotations.

\section{Experimental Results}
To evaluate the effectiveness of UAVTwin, we generate digital twins by reconstructing 3D scenes from UAV image collections. Our experiments are designed to evaluate two core components of our system: (1) the neural rendering module's ability to generate novel-view images and (2) the effectiveness of the synthesized dataset in enhancing UAV perception performance.

\subsection{Datasets and Implementation Details}
\noindent\textbf{Okutama-Action:} The Okutama-Action dataset \cite{barekatain2017okutama} comprises UAV-captured video footage within a baseball stadium, depicting multiple human agents performing a range of single-agent and multi-agent actions. The dataset is collected using two UAVs, with altitudes ranging from 10 to 45 meters and camera angles set at either 45 or 90 degrees. They collected 25 scenarios under two different lighting conditions (sunny and cloudy) and at two time periods (morning and noon). Both time and lighting differences result in significant appearance variations. Additionally, each scenario covers only a limited area due to the constraints of a single sensor and limited power.

\noindent\textbf{DroneSplat:} The DroneSplat dataset \cite{DroneSplat} consists of six scenes captured by a single UAV, including dynamic elements such as humans, cars, and buses, commonly found in urban environments. For testing, they provide the novel-view scenes without dynamic objects.

\noindent\textbf{NeRF-OSR:} The NeRF-OSR dataset \cite{NeRF-OSR} is designed for outdoor scene relighting. Similar to the NeRF-MS \cite{nerfms}, we select three sequences under different lighting conditions for each of the three scenes. For testing, we sampled every eighth frame.

\subsection{Background Rendering}
For evaluating the background rendering, we compare our method with four different methods. GOF \cite{yu2024gaussian}, 2DGS \cite{huang20242d}, and PGSR \cite{chen2024pgsr} are state-of-the-art surface reconstruction methods built on 3D Gaussian splatting.

\noindent\textbf{Comparison on Okutama-Action Dataset:} In Table \ref{table:rendering_quantitative_comparison_okutama}, we report the rendering quality using the standard PSNR, SSIM \cite{wang2004image}, and LPIPS \cite{zhang2018unreasonable} on UAV scenes in the Okutama-Action dataset. 
We calculate the average rendering metrics across 7 sequences for Drone1-Morning, 7 sequences for Drone2-Morning, 6 sequences for Drone1-Noon, and 5 sequences for Drone2-Noon. 
Each sequence contains approximately 100 to 200 images. 
% More detailed results are provided in the supplementary material. 
When applying masked loss during the training of 2DGS, we observe that the training process fails. 
Thus, except for 2DGS, we use the same refined mask for training for fair comparison. 
Since some test images include dynamic objects, we use masks for measuring metrics. 
Our method demonstrates better rendering quality compared to other recent methods such as 3DGS \cite{kerbl20233d} (with a 1.89 PSNR increase), GOF \cite{yu2024gaussian} (with a 2.44 PSNR increase), 2DGS \cite{huang20242d} (with a 1.23 PSNR increase), and PGSR \cite{chen2024pgsr} (with a 2.79 PSNR increase). 

In Fig. \ref{fig:rendering_surface_comparison}, the first row presents the shading meshes, which provide the most effective way to highlight geometric differences. Our method produces a smooth surface on the ground, free from distortions commonly observed in other approaches. Notably, although both PGSR and our method incorporate multi-view geometric regularization from Sec. \ref{sec:3.4}, sequence embedding plays a crucial role in eliminating floaters and geometric artifacts in the mesh. Rows 2–4 in Fig.\ref{fig:rendering_surface_comparison} present the rendering results. The red dotted boxes highlight regions where our method accurately captures the appearance, maintaining a color tone similar to the original scene. The yellow arrows indicate areas where strong mesh artifacts are prevalent in most algorithms, as well as rendering artifacts visible in rows 2 and 3 in Fig.\ref{fig:rendering_surface_comparison}.

%%%%%%%% Okutama-Action
\begin{table}[t]
    \centering
    \begin{adjustbox}{width=0.99\linewidth,center}
    \begin{tabular}{l|c|ccc|ccc}
    \toprule
    \multirow{2}{*}{Train}&\multirow{2}{*}{Test}& \multicolumn{3}{c|}{$\text{mAP}_{50}$} &\multicolumn{3}{c}{$\text{mAP}_{50:95}$}\\
    & & yolov8n & yolov8s & yolov8m & yolov8n & yolov8s & yolov8m \\
    \midrule
    Syn-N & Noon & 40.8 & 39.7 & 44.2 & 11.6 & 10.1 & 11.6 \\
    \midrule
    R1.2.2 & Noon & 46.8 & 52.5 & 58.0  & 14.0 & \textbf{17.0} & 18.9  \\
    R1.2.2 + Syn-N & Noon & \textbf{50.2} & \textbf{53.4} & \textbf{61.5} & \textbf{14.6} & 15.0 & \textbf{20.3} \\ 
    \midrule
    R2.2.2 & Noon & 40.7 & 43.1 & 51.6 & 12.5 & 14.1 & 16.4 \\
    R2.2.2 + Syn-N & Noon & \textbf{46.3} & \textbf{47.9} & \textbf{56.1} & \textbf{13.7} & \textbf{16.5} & \textbf{17.4} \\
    \midrule
    Syn-M & Morning & 19.7 & 16.8 & 31.3 & 4.5 & 3.7 & 21.9 \\
    \midrule
    R1.1.1 & Morning & 40.6 & 44.9 & 60.0 & 11.8 & 13.4 & 19.1 \\
    R1.1.1 + Syn-M & Morning & \textbf{43.1} & \textbf{48.1} & \textbf{61.5} & \textbf{12.5} & \textbf{14.8} & \textbf{20.3} \\ 
    \midrule
    R2.1.1 & Morning & 30.6 & 33.6 & 35.2 & \textbf{7.5} & 8.4 & 8.3\\
    R2.1.1 + Syn-M & Morning & \textbf{31.7} & \textbf{44.6} & \textbf{47.6} & 7.4 & \textbf{12.0} & \textbf{12.5} \\
    \bottomrule
    \end{tabular}
    \end{adjustbox}
    \vspace{-2mm}
    \caption{\textbf{AP Comparison on the Okutama-Action of YOLOv8 family.} ``yolov8n", ``yolov8s", and ``yolov8m" represent three YOLOv8 models \cite{yolov8} with different architecture. ``R" denotes a finetuning using real dataset. ``Syn-" denotes synthetic data used for training. ``Syn-N" refers to synthetic data in the Noon style, while ``Syn-M" represents synthetic data in the Morning style. Compared to YOLO trained with ``Real'' dataset, we improve the mAP$_{50}$ metric by 2.5 \% - 13.7 \%.   
    }
    \vspace{-2mm}
    \label{table:Perception}
\end{table}

\begin{table}[t]
    \centering
    \begin{adjustbox}{width=0.99\linewidth,center}
    \begin{tabular}{l|cccc}
    \toprule
    % \multirow{2}{*}{Method}&\multirow{2}{*}{Mesh}&\multicolumn{4}{c}{PSNR \(\uparrow\)  / SSIM \(\uparrow\)  / LPIPS \(\downarrow\) }\\
    Method & stjohann & lwp & st & Mean  \\
    \midrule
    3DGS \cite{kerbl20233d} & 13.14/ 0.571 / 0.372   & 14.41 / 0.515 / 0.406 & 13.47 / 0.512 / 0.436 & 13.67 / 0.531 / 0.404   \\
    % mip-GS \cite{yu2024mip} & X & 00.0 / 0.00 / 0.00   & 0.00 / 0.00 / 0.00 & 0.00 / 0.00 / 0.00 & 0.00 / 0.00 / 0.00 \\
    % \midrule
    GOF \cite{yu2024gaussian} & 11.67 / 0.542 / 0.411   & 13.90 / 0.527 / 0.420 & 13.18 / 0.528 / 0.428 & 12.91 / 0.528 / 0.428 \\ 
    2DGS \cite{huang20242d} & 10.77 / 0.521 / 0.415   & 14.31 / 0.530 / 0.411 & 14.31 / 0.569 / 0.420 & 13.13 / 0.54 / 0.415 \\ 
    PGSR \cite{chen2024pgsr} & 11.52 / 0.544 / 0.407   & 14.26 / 0.494 / 0.420 & 14.17 / 0.504 / 0.407 & 13.31 / 0.514 / 0.411 \\ 
    % \textcolor{red}{Lumi} \cite{} & 00.0 / 0.00 / 0.00   & 0.00 / 0.00 / 0.00 & 0.00 / 0.00 / 0.00 & 0.00 / 0.00 / 0.00 \\
    \midrule
    MsGS (Ours) & \textbf{19.05} / \textbf{0.766} / \textbf{0.222}   & \textbf{16.92} / \textbf{0.638} / \textbf{0.324} & \textbf{18.11} / \textbf{0.650} / \textbf{0.311} & \textbf{18.03} / \textbf{0.684} / \textbf{0.285} \\ 
    \bottomrule
    \end{tabular}
    \end{adjustbox}
    \vspace{-2mm}
    \caption{\textbf{Quantitative Comparison on NeRF-OSR Dataset.} We report PSNR \(\uparrow\), SSIM \(\uparrow\), and LPIPS \(\downarrow\).   
    }
    \vspace{-6mm}
    \label{table:NeRF-OSR}
\end{table}

\begin{figure}[t]
    \centering
    \includegraphics[width=0.99\linewidth]{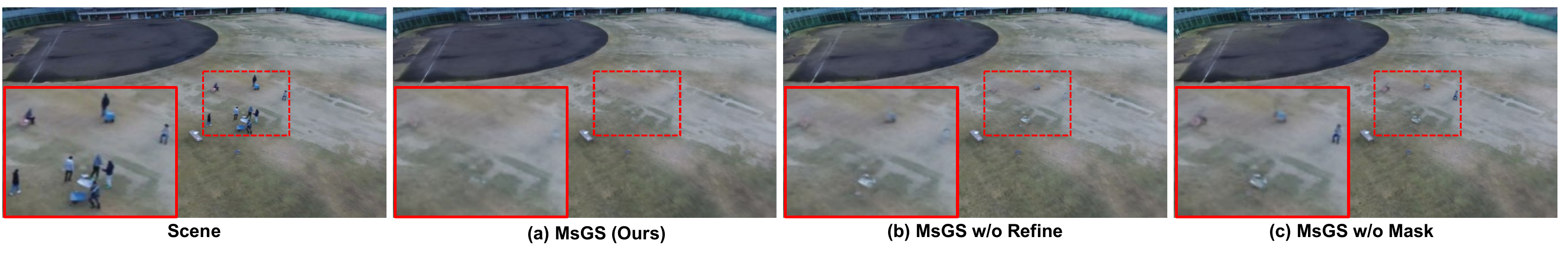}
    \vspace{-3mm}
    \caption{\textbf{Qualitative Comparison for Ablation Study}}
    \vspace{-7mm}
    \label{fig:mask_refinement_okutama}
\end{figure}

\noindent\textbf{Ablation Study:} 
Table \ref{table:rendering_quantitative_comparison_okutama} first presents the rendering results of PGSR and our method without utilizing masks for computing the training loss. 
Overall, the results indicate a slight improvement in rendering quality when masks are used. In Fig. \ref{fig:mask_refinement_okutama}, we visualize the example results. 
Additionally, ``MsGS w/o Refine" refers to training only using SAM masks conducted without the mask refinement module, which also demonstrates a minor enhancement in rendering quality.   
To further validate the components of MsGS and mask refinement, we conducted additional experiments on different datasets.
% , including DroneSplat and NeRF-OSR, which specifically address these issues.

\noindent\textbf{Appearance Modeling on NeRF-OSR Data} 
Since the official split details for NeRF-MS \cite{nerfms} are not available, we selected three sequences based on their paper and retrained all comparative algorithms for evaluation. To assess the effectiveness of our method in handling large appearance variations, we assigned three per-sequence embedding vectors to each sequence. Table \ref{table:NeRF-OSR} demonstrates that our method is robust to significant appearance variations, achieving superior performance compared to other approaches.

\noindent\textbf{Mask Refinement on DroneSplat Dataset:}  We use this dataset to evaluate the effectiveness of our neural rendering quality in handling dynamic objects. 
In Table \ref{table:DroneSplat}, we assess the validity of the mask refinement module. 
For a fair comparison, we do not use the appearance embedding vector in this experiment. 
The reported metrics are averaged across six scenes. Our method, incorporating refined masks, demonstrates slightly better performance compared to other approaches, including our method using only SAM masks (``MsGS w/o Refine"). 
Please refer to the supplementary materials for more detailed results.

\begin{table}[t]
    \centering
    \begin{adjustbox}{width=0.8\linewidth,center}
    \begin{tabular}{l|c|c|c}
    \toprule
    % \multirow{2}{*}{Method}&\multirow{2}{*}{Mesh}&\multirow{2}{*}{Mask}&\multicolumn{1}{c}{Mean}\\
    % & & & PSNR \(\uparrow\)  / SSIM \(\uparrow\)  / LPIPS \(\downarrow\)  \\
    Method & Mesh & Mask & Mean \\
    \midrule
    3DGS \cite{kerbl20233d}  & X & O & 19.51 / 0.614 / 0.295 \\
    2DGS \cite{huang20242d} & O & O & 19.45 / 0.622 / 0.282   \\ 
    PGSR \cite{chen2024pgsr} & O & O & 19.10 / \underline{0.630} / 0.284 \\ 
    \midrule
    MsGS (Ours) & O & O & \textbf{19.70} / \textbf{0.643} / \textbf{0.267} \\
    MsGS w/o Refine & O & O & \underline{19.51} / 0.625 / \underline{0.267}  \\
    \bottomrule
    \end{tabular}
    \end{adjustbox}
    % \vspace{-2mm}
    \caption{\textbf{Quantitative Comparison on DroneSplat Dataset.} We report PSNR \(\uparrow\), SSIM \(\uparrow\), and LPIPS \(\downarrow\). 
    The best and second best-performing algorithms for each metric are bolded and underlined.
    }
    % \vspace{-5mm}
    \label{table:DroneSplat}
\end{table}

\subsection{Training UAV Human Recognition}
We evaluate our generated dataset for person detection in UAV applications. In our experiments, 
we use YOLOv8 \cite{yolov8} with three different architectures. 
Each detection model is trained on a different dataset type. 
``Syn-N" and ``Syn-M" represent synthetic data in the Noon and Morning styles, respectively. 
Since we have per-sequence embedding vectors for each of the 25 sequences, we can select the corresponding sequence embedding to generate data with visually similar characteristics. 
Figure \ref{fig:data_generation_result} illustrates synthetic data that closely matches a predefined sequence. 
In Table \ref{table:Perception}, ``R1.2.2'', ``R2.2.2'', ``R1.1.1'', and ``R2.1.1'' denote real data from the Noon1.2.2, Noon2.2.2, Morning1.1.1, and Morning2.1.1 sequences, respectively. 
We consistently observe that incorporating synthetic data into real data for fine-tuning YOLOv8 improves performance in both $\text{mAP}{50}$ and $\text{mAP}{50:95}$. 
Additionally, training exclusively on synthetic data yields some detection capability, though its performance remains suboptimal.

% \subsection{Component Results}
% \label{sec:5.4}

\section{Conclusion, Limitations, and Future Work}
This paper introduces UAVTwin, a framework for data generation to train UAV-based human recognition algorithms by constructing a photorealistic digital twin using 3D Gaussian splatting. To promote photorealistic rendering, we propose MsGS for background rendering, which accommodates multi-sequence images with large appearance variations and dynamic objects. Additionally, we present a neural data generation pipeline that integrates synthetic humans with background rendering, aiming to minimize the domain gap. However, our synthesized data still suffer from a domain gap, as synthetic humans visually differ from real humans, making it challenging to achieve significant improvements in perception performance. In the future, we will explore more carefully designed object insertion techniques or utilize realistic-looking human avatars reconstructed from real-human datasets to further reduce the domain gap.

\appendix

\section{Overview of Appendix}
In this supplementary material, we present detailed experimental results and additional explanations of the manuscript, highlighting the performance of our method compared to existing approaches.

\section{Multi-view Regularization}
For multi-view regularization, we follow PGSR approach \cite{chen2024pgsr} which shows the state-of-the-art surface reconstruction performance on benchmark datasets. A patch-based normalized cross-correlation (NCC) loss is applied between two gray renders $I$ and $\hat{I}$ to force the multi-view photometric consistency. Also, the multi-view geometric consistency regularization $L_{mvegeo}$ is defined using the forward and backward projection error $\phi$ of pixel $p$ computed by using the reference frame and neighboring frame.  
\begin{equation} \label{eq:multiviewloss}
\begin{split}
L_{mvgeo} &= \sum_{p\in P} \phi(p) \\ 
L_{mvpho} &= \sum_{\text{p}\in P_{r}}\sum_{p \in \text{p}}(1 - NCC(\hat{I}(\mathcal{H}p), I(p))) 
\end{split}
\end{equation}
where $P_{r}$ is the set of all patches obtained from the rendered image $I$ and $\mathcal{H}$ is the homography matrix between the reference and neighboring frames.
The single-view normal loss is defined as:  
\begin{equation} \label{eq:multiviewloss}
\begin{split}
L_{svgeo} &= \sum_{p \in C} || N_{d}(p) - N(p)||
\end{split}
\end{equation}
where $C$ represents the rendered image and $N_{d}$ denotes the normal of the local plane. Here, four neighboring pixels are projected into 3D points, and the normal is computed using these four 3D points.
The multi-view regularization loss is defined as:
\begin{equation} \label{eq:multiviewloss}
\begin{split}
L_{mvreg} &= \lambda_{a}L_{mvgeo} + \lambda_{b}L_{mvpho} + \lambda_{c}L_{svgeo}
\end{split}
\end{equation}
where $\lambda_{a}$, $\lambda_{b}$, $\lambda_{c}$ are set to 0.01, 0.2, and 0.05.

\section{Gaussian Splatting in the Wild}
As our focus has been on building digital twins, which require rendering and surface reconstruction, we have not extensively explored prior research \cite{kulhanek2024wildgaussians,zhang2025gaussian} designed for handling large appearance variations. Additionally, for data generation, accessing the full test image or half of the test images, as typically required in this setting, is challenging.
In table \ref{table:GS-W}, we compare our method with Gaussian splatting in the wild (GS-W) \cite{zhang2025gaussian}, which is the SOTA method. 
GS-W \cite{zhang2025gaussian} surpasses MsGS (our method) in rendering quality. 
However, its evaluation requires access to the entire set of test images, as it derives appearance embeddings from the complete test image. This requirement poses challenges for integration into our data generation pipeline. 
Given this method's superior rendering quality, it would be ideal if we could interpolate its appearance embeddings and generate novel images without needing access to the test images.

\begin{table}[t]
    \centering
    \begin{adjustbox}{width=0.8\linewidth,center}
    \begin{tabular}{l|c|c|c}
    \toprule
    \multirow{2}{*}{Method}&\multirow{2}{*}{Mesh}&\multirow{2}{*}{Mask}&\multicolumn{1}{c}{Mean}\\
    & & & PSNR \(\uparrow\)  / SSIM \(\uparrow\)  / LPIPS \(\downarrow\)  \\
    \midrule
    MsGS (Ours) & O & O &  31.79 / 0.932 / 0.162 \\ 
    GS-W \cite{zhang2025gaussian} & X & X & 32.98 / 0.944 / 0.142  \\
    \bottomrule
    \end{tabular}
    \end{adjustbox}
    \vspace{-2mm}
    \caption{\textbf{Quantitative Comparison on Okutama-Action Dataset.} GS-W \cite{zhang2025gaussian} outperforms MsGS in rendering quality but requires full test image access for evaluation, making integration into our data pipeline challenging.
    }
    % \vspace{-3mm}
    \label{table:GS-W}
\end{table}

\section{Experiment Results on Archangel Dataset}
\begin{figure}[t]
    \centering
    \includegraphics[width=0.99\linewidth]{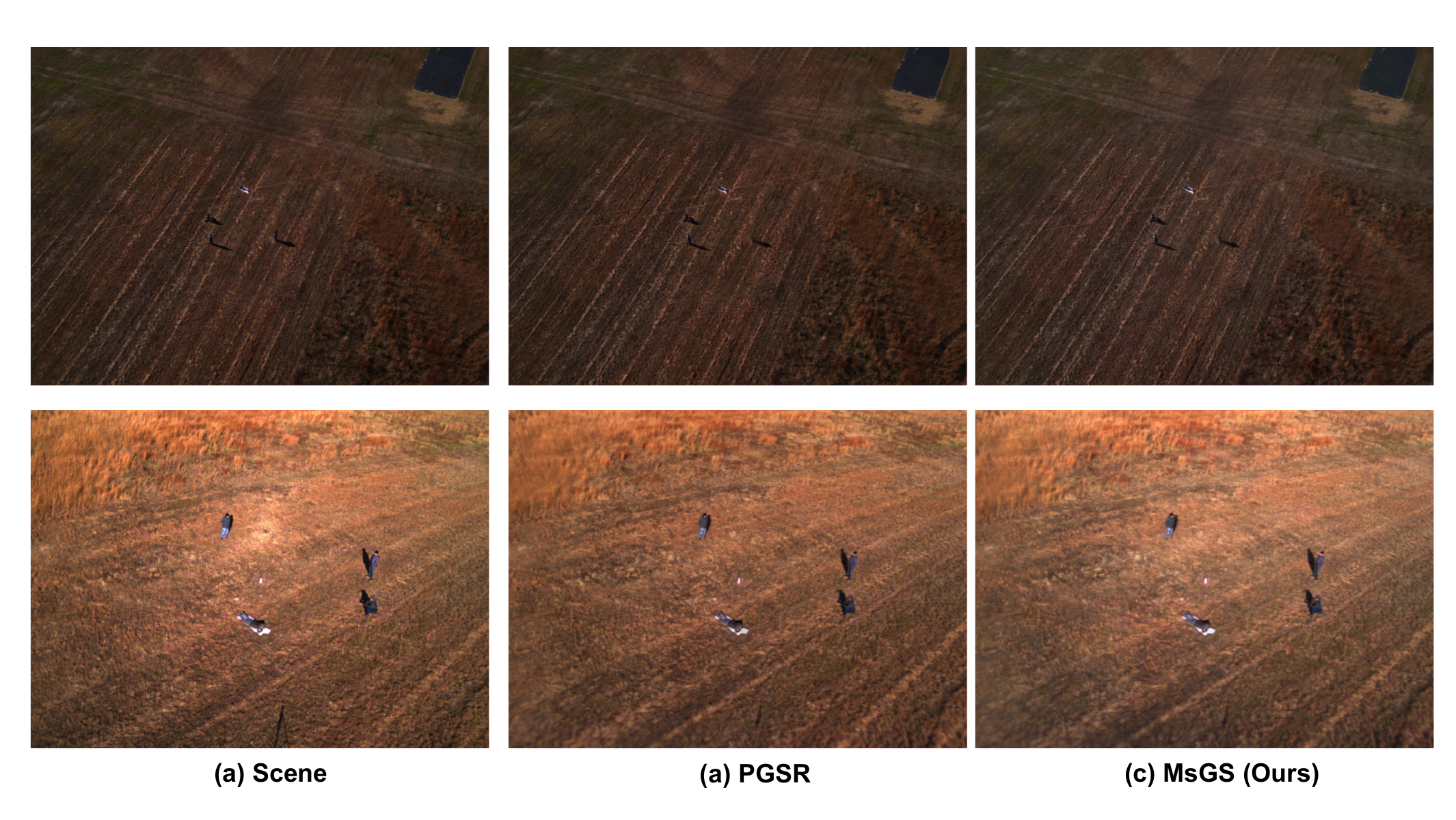}
    \vspace{-3mm}
    \caption{\textbf{Qualitative Comparison of MsGS (Our method) and PGSR \cite{chen2024pgsr}} on the Archangel dataset \cite{shen2023archangel}. }
    \label{fig:render_comparison_archangel}
\end{figure}

In Fig. \ref{fig:render_comparison_archangel}, we have applied our method to different UAV dataset, Archangel \cite{shen2023archangel}. Since the image collections are captured at different altitudes, we can generate data for a range of altitudes.

\begin{figure*}[t]
\captionsetup[subfigure]{labelformat=empty}
\centering
\begin{subfigure}{0.3\linewidth} % 0.6
  \centering
\includegraphics[width=1\linewidth]{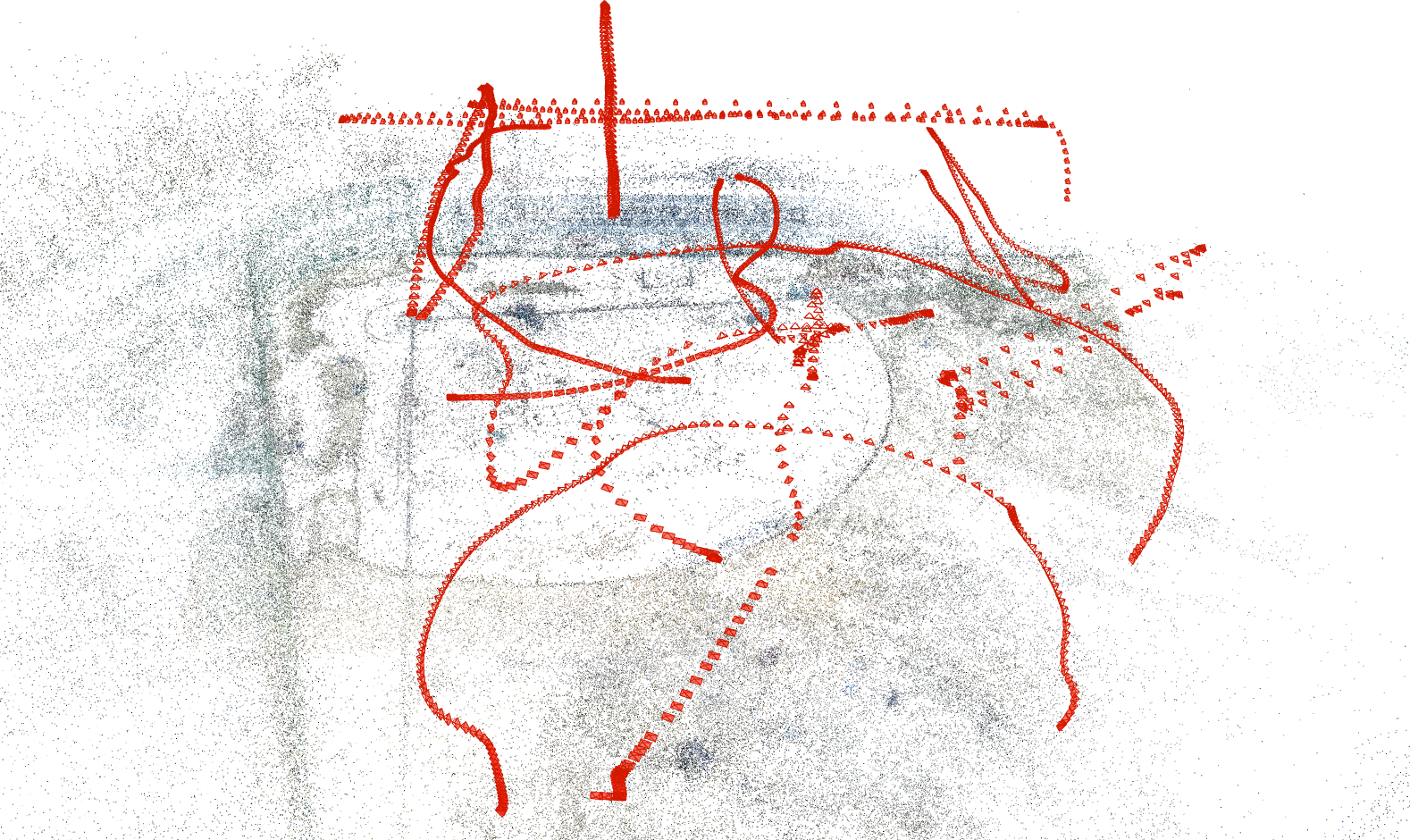}
  \caption{(a) Drone1-Noon and Drone2-Noon}
\end{subfigure}
\begin{subfigure}{0.3\linewidth} % 0.32
% \hfill
  \centering
  \includegraphics[width=\linewidth]{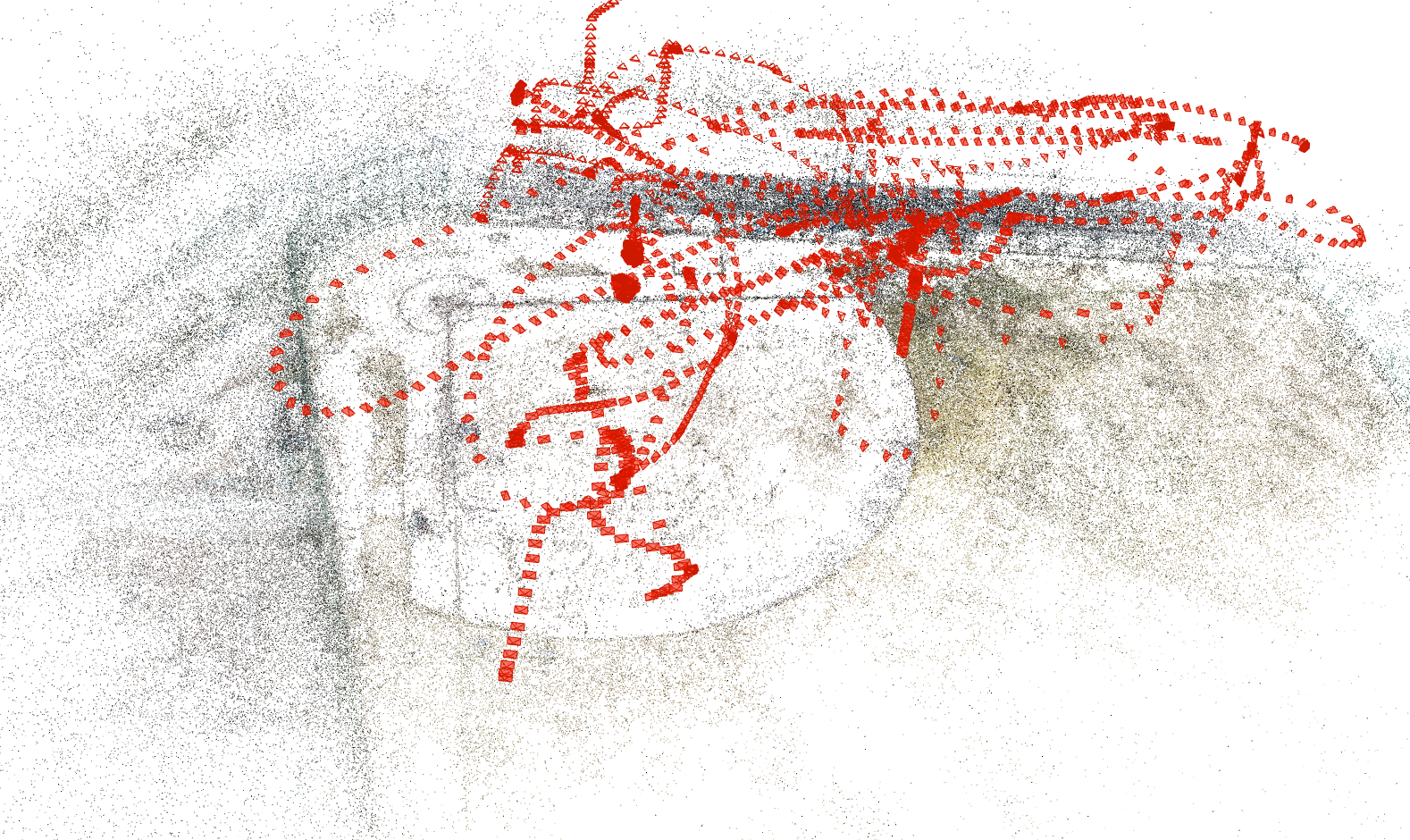}\\
  \caption{(b) Drone1-Morning and Drone2-Morning}
\end{subfigure}
\vspace{-3mm}
\caption{\textbf{Drone Camera Trajectory for Training MsGS}
(a) illustrates the UAV trajectories for Drone1-Noon and Drone2-Noon across all 11 video sequences.
(b) visualizes the UAV trajectories for Drone1-Morning and Drone2-Morning across 14 video sequences.
A single drone covers only a small area with a limited field of view, making it challenging to extrapolate neural rendering for novel camera trajectories. To address this limitation, we use multiple devices to collect a broader dataset.} 
% \vspace{-6mm}
\label{fig:colmapforall}
\end{figure*}

% \begin{figure*}[t]
% \captionsetup[subfigure]{labelformat=empty}
% \centering
% \begin{subfigure}{0.3\linewidth} % 0.6
%   \centering
% \includegraphics[width=1\linewidth]{supplementary/Scenario2_colmap.png}
%   \caption{(a) Drone1-Noon and Drone2-Noon}
% \end{subfigure}
% \begin{subfigure}{0.3\linewidth} % 0.32
% % \hfill
%   \centering
%   \includegraphics[width=\linewidth]{supplementary/Scenario3_colmap.png}\\
%   \caption{(b) Drone1-Morning and Drone2-Morning}
% \end{subfigure}
% \vspace{-6mm}
% \caption{\textbf{Drone Camera Trajectory for Training MsGS}
% (a) illustrates the UAV trajectories for Drone1-Noon and Drone2-Noon across all 11 video sequences.
% (b) visualizes the UAV trajectories for Drone1-Morning and Drone2-Morning across 14 video sequences.
% A single drone covers only a small area with a limited field of view, making it challenging to extrapolate neural rendering for novel camera trajectories. To address this limitation, we use multiple devices to collect a broader dataset.} 
% % \vspace{-6mm}
% \label{fig:colmapforall}
% \end{figure*}

\section{Implementation Details}
For building digital twin part, we implement MsGS based on Gaussian splatting \cite{kerbl20233d}.  
We use 32 dimension vector for per-sequence embedding. Thus, for 25 video sequences, we define 25 vectors for each embedding. 
For the appearance MLP, we employ two hidden layers of size 128 with ReLU activation. 
The model and Gaussian splats are optimized using the Adam optimizer \cite{kingma2014adam}. 
In mask refinement module, $\rho_1$ and $\rho_2$ are set to filter out 70 \% and 80 \% of the masks, respectively.
The appearance threshold $\rho_{pho}$ is defined based on the statistical properties of the photometric loss $L_{pho}$ values. Specifically, the mean and standard deviation of the photometric loss are computed. Instead of using a threshold set at the mean plus the standard deviation (\textit{mean} + \textit{std}), the threshold is determined as:
$\rho_{pho}$ = \textit{mean} - $\frac{std}{2}$. This formulation sets the threshold slightly below the mean, thereby reducing the influence of higher variance. As a result, values below this threshold may be considered less significant or filtered out.
We run all experiments for 30k iterations.
We set the loss weights to $\lambda_{pho}$=0.2 and $\lambda_{s}$=100.

In neural data generation, we define the human scale based on the reconstructed background mesh to control the size of synthetic humans. Specifically, we use a scale of 0.135 for the Noon scene and 0.195 for the Morning scene in the Okutama dataset. For rendering foreground humans, we utilize the Synbody rendering toolbox. We usually randomly position 10 $\sim$ 15 synthetic humans for data generation. We manually select a training image that represents the environmental lighting and input it into DiffusionLight \cite{phongthawee2024diffusionlight} to generate environmental maps. We follow the official YOLOv8 \cite{yolov8} fine-tuning protocols for training the human detection model.

\section{Mask Refinement}
\begin{figure*}[t]
    \centering
    \includegraphics[width=0.9\linewidth]{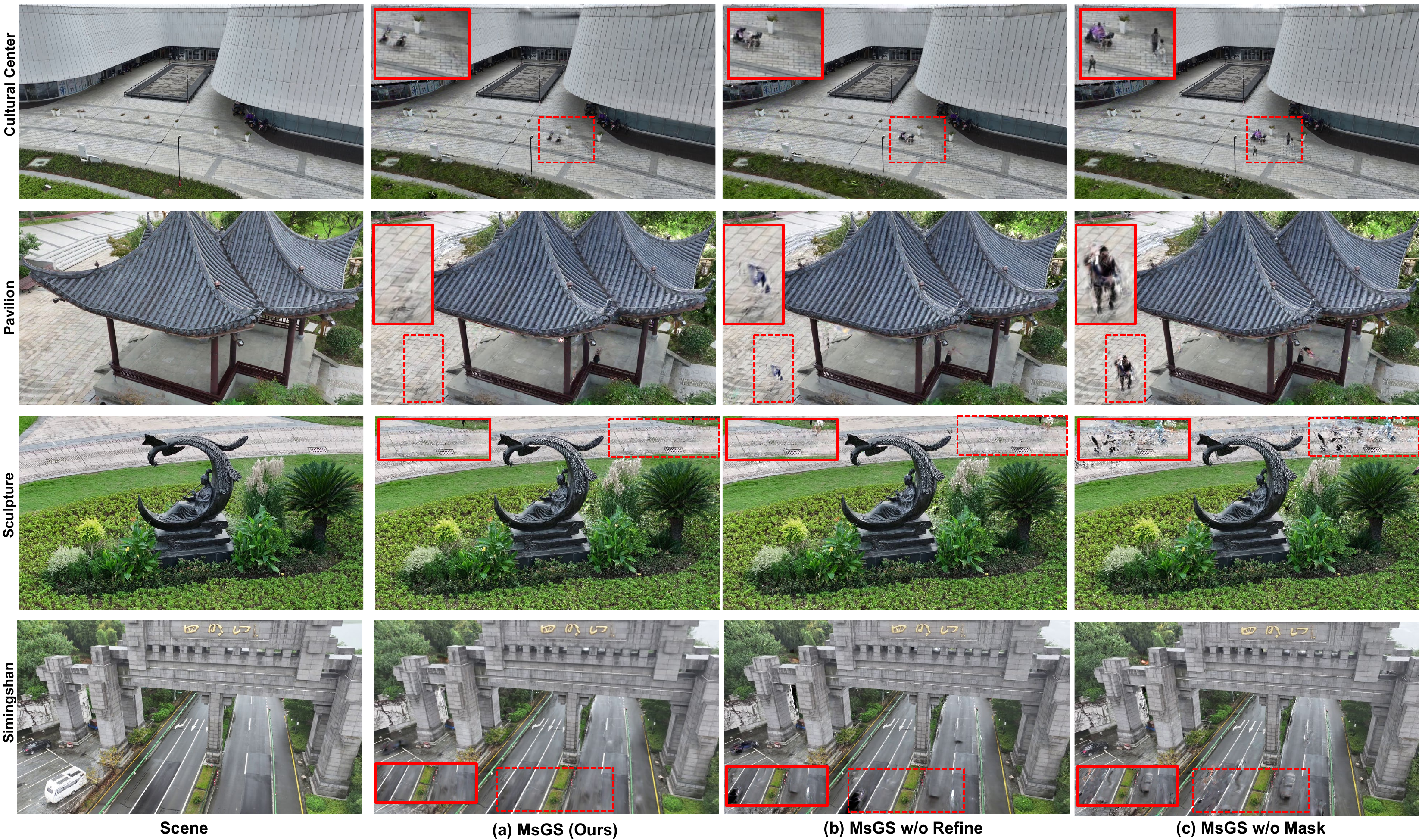}
    \vspace{-3mm}
    \caption{\textbf{Qualitative Comparison of MsGS (Ours), ``MsGS w/o Refine", ``MsGS w/o Mask"} on the DroneSplat dataset. Without masks, the 4th columns shows that significant artifacts appear in regions occupied by transient and dynamic objects. When using masks refined by the mask refinement module, our method (2nd column) effectively reduces artifacts compared to ``MsGS w/o Refine'' (3rd column), which relies solely on SAM masks.}
    \vspace{-4mm}
    \label{fig:mask_refinement_dronesplat}
\end{figure*}

For clarity, we provide a high-level overview of our mask refinement module using pseudocode in Algorithm \ref{alg:sam_filtering}. Figure \ref{fig:mask_refinement_dronesplat} illustrates the effectiveness of our method compared to using only SAM masks and our approach without masks.
%%% Mask Refinement 
\begin{algorithm}[h]
\caption{Filtering and Processing SAM Masks}
\label{alg:sam_filtering}
\begin{algorithmic}[1]
    \Require Area of each entity mask $A(S(e_{id}))$, Area of original SAM mask $A(\overline{M})$, Thresholds $\rho_1, \rho_2$, Dilation function $Dilate(\cdot)$, Error map per mask $R(e_{id})$, photometric loss threshold $\rho_{pho}$
    \Ensure Processed SAM mask $\overline{M}$
    
    \State \textbf{Thresholding:} Select entity masks $S'(e_{id})$ where $A(S(e_{id})) < \rho_1$ and $A(\overline{M}) < \rho_2$
    \State \textbf{Dilation:} Apply dilation to SAM mask: $\overline{M} \gets Dilate(\overline{M})$
    \State \textbf{Mask Selection:} Gather all entity masks that overlap with $\overline{M}$: $M' \gets Dilate(\overline{M}) \cup \hat{M}$
    
    \For{each entity mask $\hat{M}(e_{id}) \in M'$}
        \If{$R(e_{id}) > \rho_{pho}$} 
            \State Add $\hat{M}(e_{id})$ to $\overline{M}$
        \Else
            \State Remove $\hat{M}(e_{id})$ from $\overline{M}$
        \EndIf
    \EndFor  
    % \Return $\overline{M}$
\end{algorithmic}
\end{algorithm}

\section{Okutama-Action Dataset}
The Okutama-Action dataset \cite{barekatain2017okutama} comprises 25 video scenarios, captured simultaneously by two drones with different configurations. To train our MsGS, we divide the dataset into two subsets based on time of day: Morning and Noon. Each time zone exhibits significant appearance variations, necessitating separate training. Consequently, Drone1-Noon and Drone2-Noon are used together for training, while Drone1-Morning and Drone2-Morning are trained simultaneously. In Fig. \ref{fig:colmapforall}, we visualize the camera trajectories for all 25 sequences. Each sequence covers only a small area with a limited field of view.

\section{Details of Quantitative Results}
In the manuscript, due to space limitations, we report the mean values across multiple scenes from the Okutama-Action and DroneSplat datasets when evaluating neural rendering quality. In this section, Table \ref{table:DroneSplat}, \ref{table:Drone1-Morning}, \ref{table:Drone1-Noon}, \ref{table:Drone2-Morning}, \ref{table:Drone2-Noon} provide detailed metric results for all individual scenes.  

\section{Appearance Modeling on NeRF-OSR Data}

\begin{figure*}[t]
    \centering
    \includegraphics[width=0.9\linewidth]{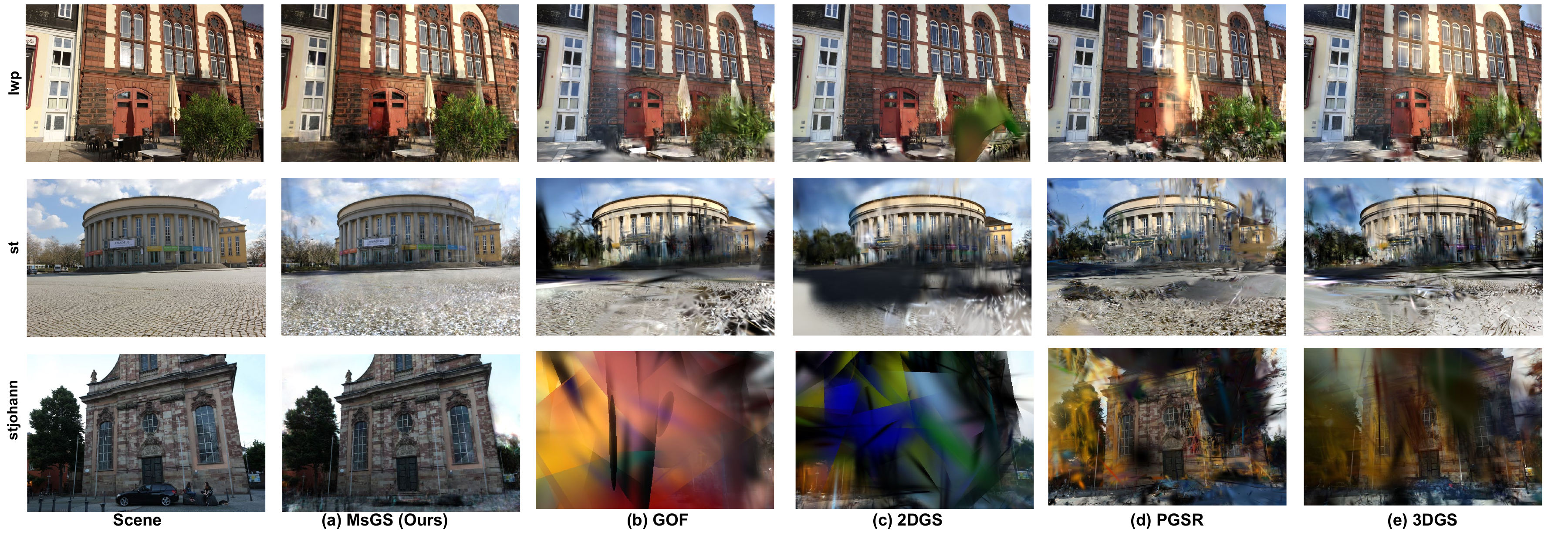}
    \vspace{-4mm}
    \caption{\textbf{Qualitative Comparison of MsGS (Our method), GOF \cite{yu2024gaussian}, 2DGS \cite{huang20242d}, PGSR \cite{chen2024pgsr}, and 3DGS \cite{kerbl20233d}} on the NeRF-OSR dataset.Due to the incorporation of per-sequence embeddings, our method exhibits strong robustness to appearance variations.}
    \vspace{-3mm}
    \label{fig:render_comparison_nerfosr}
\end{figure*}

Figure \ref{fig:render_comparison_nerfosr} visualize the effectiveness of per-sequence embedding vectors. 

\section{Visualization of Synthesized Data}
\begin{figure*}[t]
    \centering
    \includegraphics[width=0.99\linewidth]{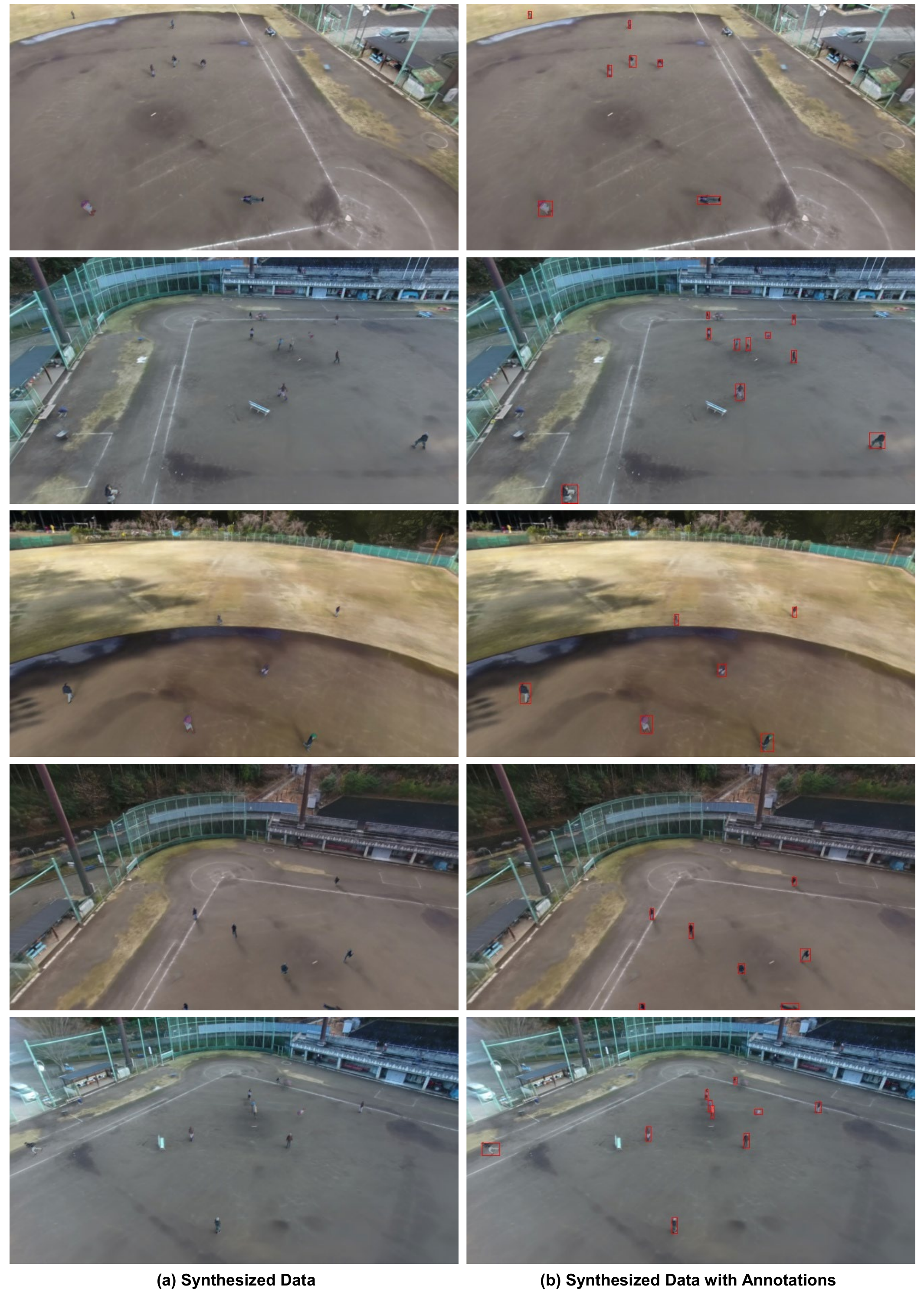}
    \vspace{-3mm}
    \caption{\textbf{Examples of Synthetized Data.} Our neural data generation provide bounding box and action label annotations.}
    % \vspace{-7mm}
    \label{fig:synth_data}
\end{figure*}

\begin{figure*}[t]
    \centering
    \includegraphics[width=0.99\linewidth]{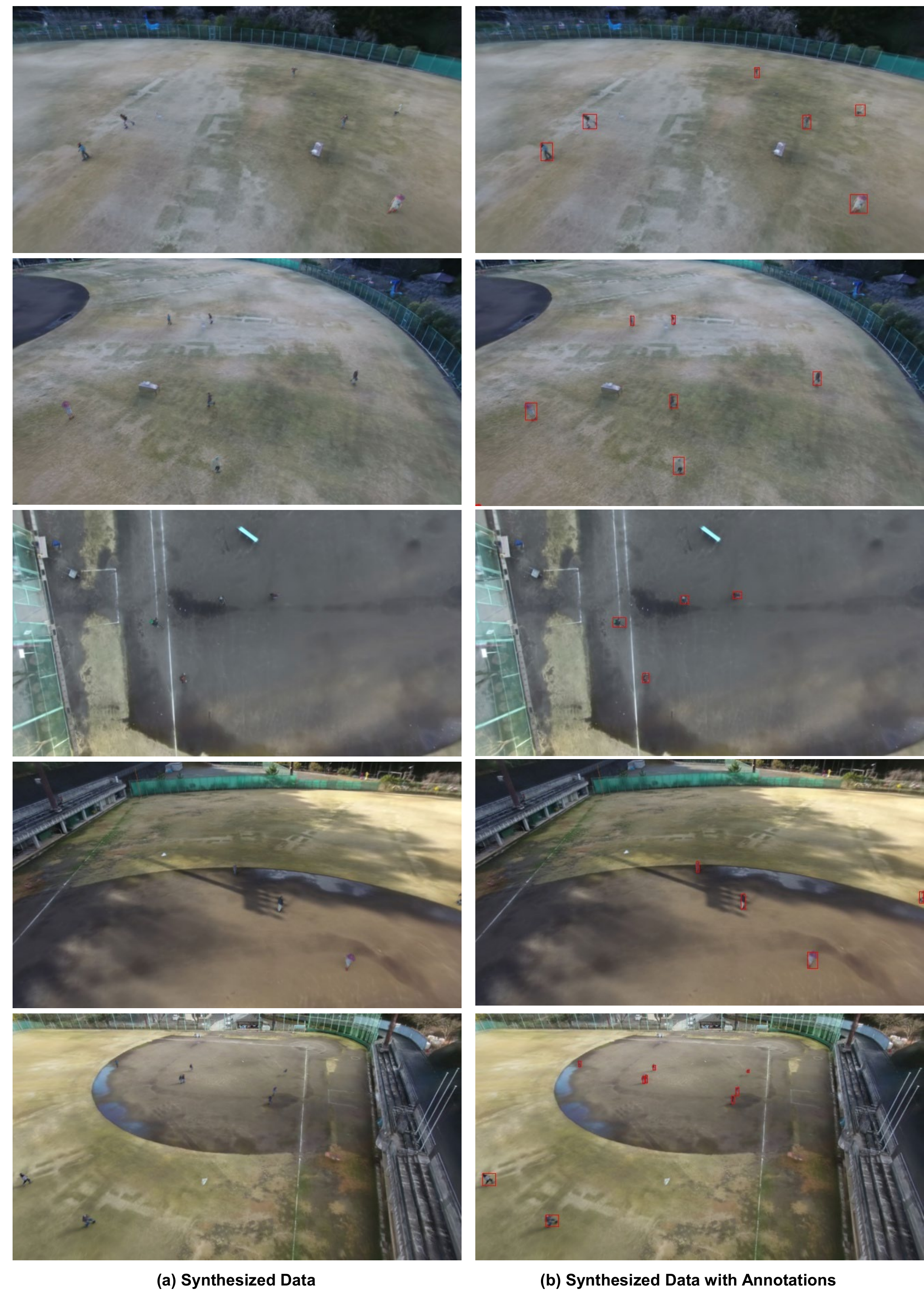}
    \vspace{-3mm}
    \caption{\textbf{Examples of Synthetized Data.} Our neural data generation provide bounding box and action label annotations.}
    % \vspace{-7mm}
    \label{fig:synth_data2}
\end{figure*}

Figures \ref{fig:synth_data} and \ref{fig:synth_data2} present examples of our synthesized data. We also showcase the generated data with overlaid bounding box annotations. Additionally, the graphics engine can produce action labels, semantic masks, depth maps, and normal maps.

\newpage

%%%%%%%% DroneSplat
\begin{table*}[t]
    \centering
    \begin{adjustbox}{width=0.99\linewidth,center}
    \begin{tabular}{l|c|c|ccccccc}
    \toprule
    \multirow{2}{*}{Method}&\multirow{2}{*}{Mesh}&\multirow{2}{*}{Mask}&\multicolumn{7}{c}{PSNR \(\uparrow\)  / SSIM \(\uparrow\)  / LPIPS \(\downarrow\) }\\
    & & & Cultural Center & Intersection & Pavilion & Sculpture & Simingshan & TangTian & Mean  \\
    \midrule
    3DGS \cite{kerbl20233d}  & X & O & 22.58 / 0.792 / 0.179   & 18.17 / 0.586 / 0.398 & 17.70 / 0.504 / 0.297 & 18.67 / 0.514 / 0.301 & 22.61 / 0.798 / 0.0.166 & 17.35 / 0.493 / 0.432 & 19.51 / 0.614 / 0.295 \\
    % mip-GS \cite{yu2024mip} & X & 00.0 / 0.00 / 0.00   & 0.00 / 0.00 / 0.00 & 0.00 / 0.00 / 0.00 & 0.00 / 0.00 / 0.00 & 0.00 / 0.00 / 0.00 \\
    % \midrule
    % GOF \cite{yu2024gaussian} & O & O & 00.0 / 0.00 / 0.00   & 0.00 / 0.00 / 0.00 & 0.00 / 0.00 / 0.00 & 0.00 / 0.00 / 0.00 & 0.00 / 0.00 / 0.00 \\ 
    2DGS \cite{huang20242d} & O & O & 21.41 / 0.712 / 0.165   & 18.57 / 0.633 / 0.379 & 17.93 / 0.535 / 0.378 & 18.90 / 0.561 / 0.255 & 22.76 / 0.898 / 0.170 & 17.13 / 0.500 / 0.431 & 19.45 / 0.622 / 0.282   \\ 
    PGSR \cite{chen2024pgsr} & O & O & 22.37 / 0.737 / 0.195   & 17.18 / 0.590 / 0.387 & 17.51 / 0.507 / 0.281 & 18.57 / 0.561 / 0.252 & 22.82 / 0.898 / 0.158 & 16.16 / 0.487 / 0.435 & 19.10 / \underline{0.630} / 0.284 \\ 
    \midrule
    MsGS (Ours) & O & O & 22.75 / 0.801 / 0.185 & 18.26 / 0.62 / 0.364 & 17.65 / 0.515 / 0.273 & 19.46 / 0.596 / 0.218 & 22.63 /  0.812 / 0.154 & 17.41 / 0.512 / 0.413 & \textbf{19.70} / \textbf{0.643} / \textbf{0.267} \\
    MsGS w/o Refine & O & O & 22.57 / 0.737 / 0.194   & 17.56 / 0.598 / 0.391 & 17.44 / 0.508 / 0.282 & 19.40 / 0.593 / 0.221 & 22.95 / 0.812 / 0.155 & 17.17 / 0.503 / 0.423 & \underline{19.51} / 0.625 / \underline{0.267}  \\
    \bottomrule
    \end{tabular}
    \end{adjustbox}
    \vspace{-2mm}
    \caption{\textbf{Quantitative Comparison on DroneSplat Dataset.} In the manuscript, we present the mean values across six scenes. This table provides detailed results for each individual scene.
    }
    \vspace{-2mm}
    \label{table:DroneSplat}
\end{table*}

%%%%%%%% scenario2
\begin{table*}[h]
    \centering
    \begin{adjustbox}{width=0.99\linewidth,center}
    \begin{tabular}{l|c|c|ccccccc}
    \toprule
    \multirow{2}{*}{Method}&\multirow{2}{*}{Mesh}&\multirow{2}{*}{Mask}&\multicolumn{7}{c}{Drone1-Noon (PSNR \(\uparrow\)  / SSIM \(\uparrow\)  / LPIPS \(\downarrow\)) } \\
    & & & 1.2.2 & 1.2.4 & 1.2.6 & 1.2.8 & 1.2.9 & 1.2.11 & Mean  \\
    \midrule
    3DGS \cite{kerbl20233d} & X & O & 32.35 / 0.945 / 0.171   & 35.80 / 0.961 / 0.143 & 24.87 / 0.826 / 0.285 & 28.50 / 0.916 / 0.166 & 31.76 / 0.895 / 0.189 & 32.39 / 0.918 / 0.158 & 31.48 / 0.917 / 0.177 \\
    % mip-GS \cite{yu2024mip} & X & 00.0 / 0.00 / 0.00   & 0.00 / 0.00 / 0.00 & 0.00 / 0.00 / 0.00 & 0.00 / 0.00 / 0.00 & 0.00 / 0.00 / 0.00 & 0.00 / 0.00 / 0.00 & 0.00 / 0.00 / 0.00 \\
    % \midrule
    GOF \cite{yu2024gaussian} & O & O & 31.29 / 0.941 / 0.170 & 34.09 / 0.957 / 0.151 & 25.07 / 0.845 / 0.274 & 26.97 / 0.886 / 0.214 & 29.72 / 0.914 / 0.185 & 30.01 / 0.912 / 0.174 & 29.95 / 0.914 / 0.189 \\ 
    2DGS \cite{huang20242d} & O & X & 30.82 / 0.934 / 0.216 & 34.28 / 0.951 / 0.194 & 24.73 / 0.868 / 0.257 & 29.32 / 0.930 / 0.159 & 33.59 / 0.921 / 0.182 & 31.63 / 0.907 / 0.202 & 31.22 / 0.923 / 0.197 \\ 
    PGSR \cite{chen2024pgsr} & O & X & 24.07 / 0.938 / 0.141 & 33.27 / 0.963 / 0.113 & 24.70 / 0.889 / 0.202 & 29.16 / 0.943 / 0.114 & 30.30 / 0.936 / 0.129 & 29.76 / 0.921 / 0.137 & 29.17 / 0.936 / 0.137 \\ 
    PGSR \cite{chen2024pgsr} & O & O & 25.19 / 0.943 / 0.140 & 33.49 / 0.963 / 0.113 & 24.49 / 0.891 / 0.200 & 29.04 / 0.942 / 0.116 & 30.76 / 0.937 / 0.129 & 30.14 / 0.920 / 0.137 & 29.47 / 0.937 / 0.137 \\ 
    \midrule
    MsGS (Ours) & O & O & 33.17 / 0.946 / 0.165 & 35.49 / 0.962 / 0.153 & 30.31 / 0.904 / 0.167 & 31.56 / 0.944 / 0.120 & 34.13 / 0.926 / 0.157 & 32.20 / 0.914 / 0.161 & 33.08 / 0.937 / 0.154 \\ 
    \bottomrule
    \end{tabular}
    \end{adjustbox}
    \vspace{-2mm}
    \caption{\textbf{Quantitative Comparison on Okutama Dataset.} Drone1-Noon comprises six video sequences, each with distinct characteristics. In the manuscript, we present the mean values across these six scenes, while this table provides detailed results for each individual scene.
    }
    \vspace{-2mm}
    \label{table:Drone1-Noon}
\end{table*}

%%%%%%%% scenario2
\begin{table*}[h]
    \centering
    \begin{adjustbox}{width=0.99\linewidth,center}
    \begin{tabular}{l|c|c|cccccc}
    \toprule
    \multirow{2}{*}{Method}&\multirow{2}{*}{Mesh}&\multirow{2}{*}{Mask}&\multicolumn{6}{c}{Drone2-Noon (PSNR \(\uparrow\)  / SSIM \(\uparrow\)  / LPIPS \(\downarrow\)) } \\
    & & & 2.2.2 & 2.2.4 & 2.2.8 & 2.2.9 & 2.2.11 & Mean  \\
    \midrule
    3DGS \cite{kerbl20233d} & X & O & 32.76 / 0.935 / 0.199 & 31.00 / 0.934 / 0.188 & 38.71 / 0.962 / 0.173 & 33.95 / 0.943 / 0.133 & 34.05 / 0.928 / 0.177 & 34.10 / 0.943 / 0.175 \\
    % mip-GS \cite{yu2024mip} & X & 00.0 / 0.00 / 0.00   & 0.00 / 0.00 / 0.00 & 0.00 / 0.00 / 0.00 & 0.00 / 0.00 / 0.00 & 0.00 / 0.00 / 0.00 & 0.00 / 0.00 / 0.00 & 0.00 / 0.00 / 0.00 \\
    % \midrule
    GOF \cite{yu2024gaussian} & O & O & 29.97 / 0.929 / 0.209 & 33.60 / 0.963 / 0.144 & 32.99 / 0.934 / 0.254 & 31.63 / 0.928 / 0.166 & 30.44 / 0.919 / 0.203 & 32.09 / 0.938 / 0.193 \\ 
    2DGS \cite{huang20242d} & O & X & 30.13 / 0.923 / 0.241 & 31.08 / 0.941 / 0.199 & 36.75 / 0.952 / 0.228 & 31.56 / 0.933 / 0.167 & 32.12 / 0.912 / 0.227 & 32.49 / 0.936 / 0.211 \\ 
    PGSR \cite{chen2024pgsr} & O & X & 31.31 / 0.943 / 0.172 & 30.04 / 0.952 / 0.141 & 34.75 / 0.968 / 0.122 & 32.78 / 0.944 / 0.131 & 29.37 / 0.924 / 0.177 & 31.94 / 0.950 / 0.143 \\ 
    PGSR \cite{chen2024pgsr} & O & O & 31.96 / 0.943 / 0.174   & 29.19 / 0.940 / 0.141 & 34.77 / 0.966 / 0.122 & 32.67 / 0.943 / 0.131 & 29.70 / 0.924 / 0.177 & 31.84 / 0.946 / 0.152 \\ 
    \midrule
    MsGS (Ours) & O & O & 31.54 / 0.935 / 0.196 & 31.89 / 0.945 / 0.167 & 37.19 / 0.959 / 0.175 & 33.69 / 0.941 / 0.123 & 33.04 / 0.921 / 0.184 & 33.59 / 0.944 / 0.188 \\ 
    \bottomrule
    \end{tabular}
    \end{adjustbox}
    \vspace{-2mm}
    \caption{\textbf{Quantitative Comparison on Okutama Dataset.} ``-" denotes a missing implementation. 
    Drone2-Noon comprises five video sequences, each with distinct characteristics. In the manuscript, we present the mean values across these six scenes, while this table provides detailed results for each individual scene.
    }
    \vspace{-2mm}
    \label{table:Drone2-Noon}
\end{table*}

%%%%%%%% scenario2
\begin{table*}[h]
    \centering
    \begin{adjustbox}{width=0.99\linewidth,center}
    \begin{tabular}{l|c|c|cccccccc}
    \toprule
    \multirow{2}{*}{Method}&\multirow{2}{*}{Mesh}&\multirow{2}{*}{Mask}&\multicolumn{8}{c}{Drone1-Morning (PSNR \(\uparrow\)  / SSIM \(\uparrow\)  / LPIPS \(\downarrow\)) } \\
    & & & 1.1.1 & 1.1.2 & 1.1.3 & 1.1.4 & 1.1.7 & 1.1.10 & 1.1.10 & Mean  \\
    \midrule
    3DGS \cite{kerbl20233d} & X & O & 28.19/ 0.926 / 0.202   & 28.70 / 0.912 / 0.223 & 27.13 / 0.917 / 0.244 & 23.35 / 0.890 / 0.220 & 25.35 / 0.881 / 0.254 & 22.83 / 0.864 / 0.220 & 29.07 / 0.915 / 0.155 & 26.33 / 0.901 / 0.222 \\
    % mip-GS \cite{yu2024mip} & X & 00.0 / 0.00 / 0.00   & 0.00 / 0.00 / 0.00 & 0.00 / 0.00 / 0.00 & 0.00 / 0.00 / 0.00 & 0.00 / 0.00 / 0.00 & 0.00 / 0.00 / 0.00 & 0.00 / 0.00 / 0.00 \\
    % \midrule
    GOF \cite{yu2024gaussian} & O & O & 29.45 / 0.937 / 0.159   & 28.29 / 0.914 / 0.183 & 30.73 / 0.928 / 0.192 & 30.96 / 0.940 / 0.087 & 29.43 / 0.910 / 0.176 & 22.43 / 0.853 / 0.233 & 29.69 / 0.914 / 0.159 & 29.04 / 0.918 / 0.164 \\ 
    2DGS \cite{huang20242d} & O & X & 28.23 / 0.924 / 0.232   & 28.24 / 0.899 / 0.258 & 27.06 / 0.914 / 0.279 & 27.55 / 0.898 / 0.204 & 27.86 / 0.881 / 0.239 & 19.57 / 0.865 / 0.247 & 27.89 / 0.916 / 0.164 & 27.09 / 0.899 / 0.236 \\ 
    PGSR \cite{chen2024pgsr} & O & X & 28.49 / 0.934 / 0.155   & 26.68 / 0.914 / 0.176 & 25.75 / 0.922 / 0.232 & 23.64 / 0.897 / 0.185 & 26.37 / 0.896 / 0.206 & 18.18 / 0.841 / 0.221 & 27.59 / 0.911 / 0.149 & 25.57 / 0.906 / 0.188 \\ 
    PGSR \cite{chen2024pgsr} & O & O & 28.77 / 0.942 / 0.135   & 27.79/ 0.919 / 0.147 & 29.54 / 0.934 / 0.160 & 30.04 / 0.939 / 0.079 & 28.03 / 0.914 / 0.151 & 18.21 / 0.842 / 0.223 & 26.93 / 0.909 / 0.154 & 27.71 / 0.920 / 0.142 \\ 
    \midrule
    MsGS (Ours) & O & O & 31.30 / 0.940 / 0.145 & 30.10 / 0.925 / 0.176 & 29.90 / 0.933 / 0.219 & 31.12 / 0.918 / 0.150 & 32.12 / 0.912 / 0.180 & 30.43 / 0.919 / 0.146 & 33.35 / 0.931 / 0.124 & 31.07 / 0.925 / 0.166 \\ 
    \bottomrule
    \end{tabular}
    \end{adjustbox}
    
    \vspace{-2mm}
    \caption{\textbf{Quantitative Comparison on Okutama Dataset.} Drone1-Morning comprises seven video sequences, each with distinct characteristics. In the manuscript, we present the mean values across these six scenes, while this table provides detailed results for each individual scene.    }
    \vspace{-2mm}
    \label{table:Drone1-Morning}
\end{table*}

%%%%%%%% scenario2
\begin{table*}[h]
    \centering
    \begin{adjustbox}{width=0.99\linewidth,center}
    \begin{tabular}{l|c|c|cccccccc}
    \toprule
    \multirow{2}{*}{Method}&\multirow{2}{*}{Mesh}&\multirow{2}{*}{Mask}&\multicolumn{8}{c}{Drone2-Morning (PSNR \(\uparrow\)  / SSIM \(\uparrow\)  / LPIPS \(\downarrow\)) } \\
    & & & 2.1.1 & 2.1.2 & 2.1.3 & 2.1.4 & 2.1.6 & 2.1.7 & 2.1.10 & Mean  \\
    \midrule
    3DGS \cite{kerbl20233d} & X & O & 29.69 / 0.915 / 0.164  & 28.92 / 0.940 / 0.148 & 30.06 / 0.920 / 0.211 & 27.82 / 0.881 / 0.233 & 31.18 / 0.923 / 0.141 & 19.73 / 0.804 / 0.377 & 25.61/ 0.888 / 0.216 & 27.32 / 0.891 / 0.216 \\
    % mip-GS \cite{yu2024mip} & X & 00.0 / 0.00 / 0.00   & 0.00 / 0.00 / 0.00 & 0.00 / 0.00 / 0.00 & 0.00 / 0.00 / 0.00 & 0.00 / 0.00 / 0.00 & 0.00 / 0.00 / 0.00 & 0.00 / 0.00 / 0.00 \\
    % \midrule
    GOF \cite{yu2024gaussian} & O & O & 29.47 / 0.907 / 0.177   & 25.48 / 0.912 / 0.229 & 26.66 / 0.911 / 0.218 & 24.13 / 0.876 / 0.160 & 30.00 / 0.914 / 0.226 & 24.32 / 0.874 / 0.222 & 25.30 / 0.884 / 0.00 & 26.43 / 0.896 / 0.212 \\ 
    2DGS \cite{huang20242d} & O & X & 29.97 / 0.910 / 0.190  & 28.07 / 0.919 / 0.241 & 31.12 / 0.925 / 0.214 & 25.83 / 0.878 / 0.260 & 30.41 / 0.918 / 0.163 & 27.51 / 0.885 / 0.247 & 24.56 / 0.877 / 0.250 & 28.33 / 0.901 / 0.222 \\ 
    PGSR \cite{chen2024pgsr} & O & X & 28.23 / 0.917 / 0.134   & 23.77 / 0.938 / 0.118 & 27.80 / 0.932 / 0.144 & 24.07 / 0.875 / 0.233 & 30.20 / 0.924 / 0.131 & 24.25 / 0.888 / 0.185 & 23.05 / 0.879 / 0.198 & 26.11 / 0.907 / 0.166 \\ 
    PGSR \cite{chen2024pgsr} & O & O & 28.41 / 0.918 / 0.132   & 24.09 / 0.940 / 0.115 & 28.10 / 0.932 / 0.141 & 24.01 / 0.874 / 0.237 & 30.44 / 0.924 / 0.131 & 28.16 / 0.909 / 0.132 & 23.17 / 0.878 / 0.199 & 26.96 / 0.911 / 0.156 \\ 
    \midrule
    MsGS (Ours) & O & O & 30.76 / 0.924 / 0.142 & 32.69 / 0.950 / 0.114 & 32.47 / 0.939 / 0.159 & 29.42 / 0.897 / 0.219 & 30.63 / 0.925 / 0.134 & 28.54 / 0.910 / 0.169 & 25.76 / 0.895 / 0.173 & 29.89 / 0.918 / 0.161 \\ 
    \bottomrule
    \end{tabular}
    \end{adjustbox}
    \vspace{-2mm}
    \caption{\textbf{Quantitative Comparison on Okutama Dataset.} Drone2-Morning comprises seven video sequences, each with distinct characteristics. In the manuscript, we present the mean values across these six scenes, while this table provides detailed results for each individual scene.
    }
    \vspace{-2mm}
    \label{table:Drone2-Morning}
\end{table*}

\newpage

% \newpage

\clearpage
\twocolumn
{
    \small
    \bibliographystyle{ieeenat_fullname}
    \bibliography{main}
}

\end{document}